\documentclass{article}
\usepackage{microtype}
\usepackage{graphicx}
\usepackage{subcaption}
\usepackage{booktabs}
\usepackage{hyperref}

\usepackage[accepted]{icml2026}

\usepackage{amsmath}
\usepackage{amssymb}
\usepackage{mathtools}
\usepackage{amsthm}
\usepackage[capitalize,noabbrev]{cleveref}
\theoremstyle{plain}

\theoremstyle{definition}

\theoremstyle{remark}

\usepackage{titletoc}

\usepackage[textsize=tiny]{todonotes}

\usepackage{float}
\usepackage{hyperref}
\usepackage{url}
\usepackage{microtype}
\usepackage{arydshln} 
\usepackage{inconsolata}
\usepackage{graphicx}
\usepackage{setspace}
\usepackage{hyperref}
\usepackage{url}
\usepackage{adjustbox} 
\usepackage{listings}
\usepackage{multirow}
\usepackage{caption}
\usepackage[most]{tcolorbox}
\usepackage{titletoc}
\usepackage[capitalize]{cleveref} 
\usepackage{float}
\usepackage{tabularx}
\usepackage{array}

\usepackage{enumitem}
\usepackage{bbold}
\usepackage[misc]{ifsym}
\usepackage{CJKutf8}
\usepackage{CJK}
\usepackage{color}
\usepackage{bbding}
\usepackage[normalem]{ulem}
\useunder{\uline}{\ul}{}
\usepackage{wrapfig}
\usepackage{siunitx}
\usepackage{makecell}
\usepackage{pifont}

\usepackage{longtable}

\usepackage{xspace}
\usepackage{booktabs,dcolumn}
\usepackage{amsmath,amsthm,amsfonts,amssymb, bm,stmaryrd,bbm}
\usepackage{colortbl}

\usepackage{titlesec}

\titlespacing{\paragraph}{%
  0pt}{%
  0.3\baselineskip}{%
  0.5em}%
\titlespacing*{\section}{0pt}{0.4\baselineskip}{0.4\baselineskip}
\titlespacing*{\subsection}{0pt}{0.3\baselineskip}{0.3\baselineskip}

\usepackage{cascadia-code}
\usepackage{courier}

\usepackage{helvet}

\definecolor{LightSteelBlue4}{RGB}{96,123,139}
\definecolor{DodgerBlue4}{RGB}{16,78,139}
\definecolor{Turquoise4}{RGB}{0,134,139}
\definecolor{Green4}{RGB}{0,139,0}
\definecolor{Brown3}{RGB}{205,85,85}
\definecolor{Azure3}{RGB}{193,205,205}
\usepackage{subcaption}
\usepackage{tikz}
\usetikzlibrary{calc}

\tikzstyle{prompt} = [rectangle,
text centered, 
minimum width = 2cm,
minimum height = 0.3cm,
font=\fontfamily{CascadiaCode-TLF}\selectfont, %
fill=LightSteelBlue4,
text=white
]

\tikzstyle{llm} = [rectangle, rounded corners,
text centered, 
draw=DodgerBlue4,
text=DodgerBlue4,
minimum width = 2cm,
minimum height = 0.6cm,
font=\small\sffamily, %
line width=1.0pt, 
fill={rgb,255:red,223;green,236;blue,248} 
]

\tikzstyle{resp} = [rectangle, %
text centered, 
draw=none,
font=\fontfamily{CascadiaCode-TLF}\selectfont,
fill=Turquoise4,
text=white
]

\tikzstyle{correct} = [rectangle, inner sep=4pt, 
text centered, 
draw=Green4,
text=Green4,
line width=1.0pt,
font=\large,
minimum width = 1.65cm,
minimum height = 0.3cm,
fill={rgb,255:red,240;green,255;blue,240}
]

\tikzstyle{wrong} = [rectangle, inner sep=4pt,
text centered, 
draw=Brown3,
text=Brown3,
line width=1.0pt,
font=\large,
minimum width = 1.65cm,
minimum height = 0.3cm,
fill={rgb,255:red,255;green,240;blue,240}
]

\tikzstyle{arrow} = [->,>=stealth,
line width=1.0pt,
draw=Azure3,
fill=Azure3
]

\tikzstyle{textlabel} = [font=\footnotesize\itshape]

\tikzstyle{sresp}=[resp,rotate=90,font=\small\fontfamily{CascadiaCode-TLF}\selectfont,inner sep=2pt]

\usepackage{xcolor}

\definecolor{bytedsa}{HTML}{315ab4}
\definecolor{bytedsb}{HTML}{3d8cff}
\definecolor{bytedsc}{HTML}{00c8d2}
\definecolor{bytedsd}{HTML}{79e6dd}

\newcommand{\task}{{SQL debugging}\xspace}
\newcommand{\ourbench}{{Squirrel Benchmark}\xspace}
\newcommand{\ourbenchsyn}{{Squirrel-Syntax}\xspace}
\newcommand{\ourbenchsem}{{Squirrel-Semantic}\xspace}

\newcommand{\ttsql}{{Text-to-SQL}\xspace}

\newtcolorbox{observationbox}[1][]{
        colback=envfill,
        colbacktitle=envfill,
        colframe=envborder,
        arc=5pt,
        fontupper=\small,
        fonttitle=\bfseries\color{black},
        boxrule=0.5mm,
        boxsep=1mm,
        width=\linewidth,
        breakable,
        title={Observation \hfill #1},
        rounded corners,
        toptitle=0.7mm,
        bottomtitle=0.7mm
}
\newtcolorbox{goldpatchbox}[1][]{
        colback=goldpatchfill,
        colbacktitle=goldpatchfill,
        colframe=goldpatchborder,
        arc=5pt,
        fontupper=\small,
        fonttitle=\bfseries\color{black},
        boxrule=0.5mm,
        boxsep=1mm,
        width=\linewidth,
        breakable,
        title={Gold Patch \hfill #1},
        rounded corners,
        toptitle=0.7mm,
        bottomtitle=0.7mm
}
\newtcolorbox{issuebox}[1][]{
        colback=issuefill,
        colbacktitle=issuefill,
        colframe=issueborder,
        arc=5pt,
        fontupper=\small,
        fonttitle=\bfseries\color{black},
        boxrule=0.5mm,
        boxsep=1mm,
        width=\linewidth,
        breakable,
        title={Issue \hfill #1},
        rounded corners,
        toptitle=1mm
}
\newtcolorbox{agentbox}[1][]{
        colback=agentfill,
        colbacktitle=agentfill,
        colframe=agentborder,
        arc=5pt,
        fontupper=\small,
        fonttitle=\bfseries\color{black},
        boxrule=0.5mm,
        boxsep=1mm,
        width=\linewidth,
        breakable,
        title={SWE-agent \hfill #1},
        rounded corners,
        toptitle=1mm,
        lower separated=false
}
\newtcolorbox{fileviewerbox}[1]{
        enhanced,
        breakable,
        boxrule = 1.5pt,
        fontupper = \small,
        fonttitle = \bf\color{black},
        arc = 5pt,
        rounded corners,
        colframe = black,
        colbacktitle = swecream,
        colback = swecream,
        title = #1,
        left=4pt 
}
\newtcolorbox{promptbox}[1]{
    enhanced,
    breakable,
    boxrule=1pt,  
    fontupper=\small,
    fonttitle=\bfseries\color{black},
    arc=3pt,  
    rounded corners,
    colframe=black,
    colbacktitle=swecream,
    colback=swecream,
    title=#1,
    left=2mm,  
    right=2mm,  
    top=1mm,  
    bottom=1mm  
}

\icmltitlerunning{Beyond Text-to-SQL: Can LLMs Really Debug Enterprise ETL SQL?}

\begin{document}

\twocolumn[
  \icmltitle{\includegraphics[height=1.5em]{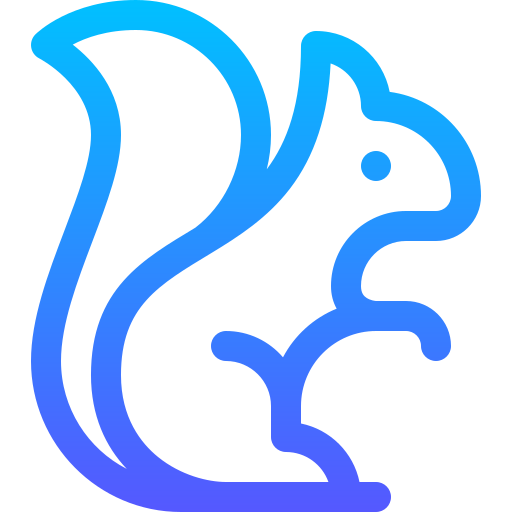}\hspace{0.2em} 
    Beyond Text-to-SQL: Can LLMs Really Debug Enterprise ETL SQL?}

  \begin{icmlauthorlist}
    \icmlauthor{Jing Ye}{2}
    \icmlauthor{Yiwen Duan}{1}
    \icmlauthor{Yonghong Yu}{1}
    \icmlauthor{Victor Ma}{2}
    \icmlauthor{Yang Gao}{1}
    \icmlauthor{Xing Chen}{1}
  \end{icmlauthorlist}

  \icmlaffiliation{1}{Bytedance Inc., Beijing, China}
  \icmlaffiliation{2}{Independent Researcher}

  \icmlcorrespondingauthor{Xing Chen}{chenxing.xc@bytedance.com}

  \icmlkeywords{Text-to-SQL, Agent, AI Coding, Benchmark}

  \vskip 0.3in
]

\printAffiliationsAndNotice{}

\begin{abstract}
SQL is central to enterprise data engineering, yet generating fully correct SQL code in a single attempt remains difficult—even for experienced developers and advanced \ttsql LLMs—often requiring multiple debugging iterations.
We introduce \textbf{\ourbench}, the first benchmark for enterprise-level SQL reasoning and debugging. Our benchmark is built upon two key innovations: (1) an \textbf{automated construction workflow} that employs reverse engineering to systematically inject realistic bugs into large-scale SQL code, enabling scalable and diverse benchmark generation; and (2) an \textbf{execution-free evaluation framework} tailored for enterprise settings, providing fast, accurate, and resource-efficient assessment.
\ourbench comprises $469$ \ourbenchsyn queries featuring syntax errors with explicit error messages, and $516$ \ourbenchsem queries targeting semantic errors where codes fails to meet user intent.
The queries are highly complex, averaging over $140$ lines, and featuring deep and wide abstract syntax trees (average width $>11$, depth $>8.7$).
Evaluation of nearly $30$ LLMs reveals a substantial performance gap: the best-performing model, Claude-4-Sonnet, achieves only $36.46\%$ accuracy on \ourbenchsyn and $32.17\%$ on \ourbenchsem, while most models score below $20\%$. We further explore four solution strategies, identify key challenges, and outline promising directions for enterprise SQL debugging with LLMs.
\end{abstract}
\section{Introduction}
Databases are a cornerstone of modern data infrastructure, powering critical applications across finance, web services, and scientific computing. Structured Query Language (SQL) remains the predominant interface for human–data interaction, enabling large-scale extraction, transformation, and loading (ETL) workflows \citep{chamberlin1974, armbrust2015}. Recent research on \ttsql large language models (LLMs) has sought to help analysts automate routine queries, streamline data workflows, and support advanced business intelligence \citep{zhong2017wikisql,yu2018spider,li2025omnisql}. 

\begin{figure*}[ht]
    \centering
    \includegraphics[width=1\textwidth]{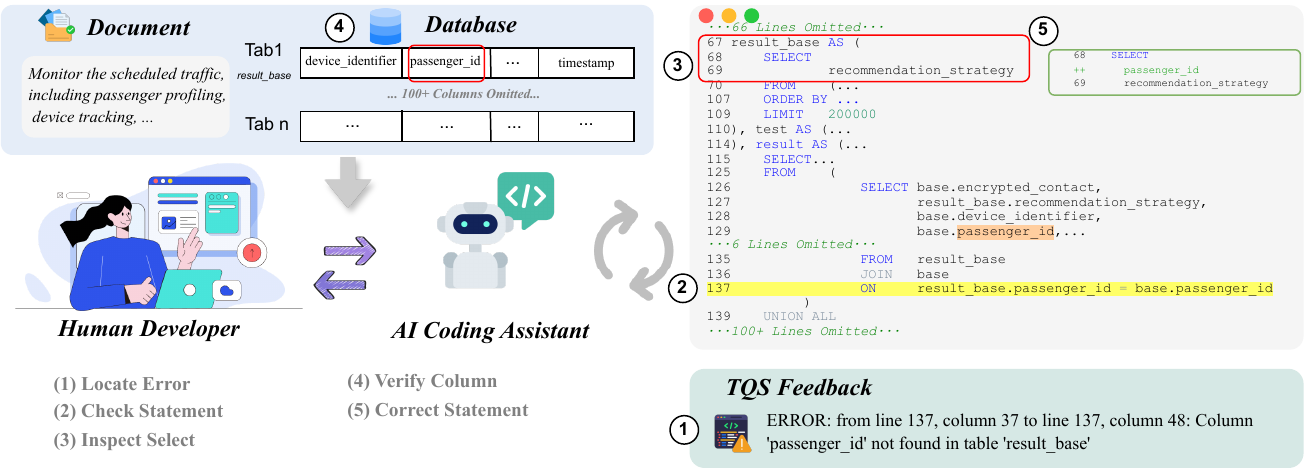}
    \caption{\ourbench evaluates LLMs on real-world enterprise-level \task workflows. 
    It involves multi-step reasoning and actions, including understanding requirements and schemas, diagnosing error messages, and iteratively refining scripts through cycles of reasoning and debugging.}
    \label{fig:task}
\end{figure*}

Enterprise SQL code is often lengthy, complex, and deeply nested, making it extremely challenging for both experienced developers and \ttsql LLMs to generate correct code in a single attempt \citep{lei2025spider}. Instead, success typically requires multi-step reasoning and iterative debugging. As shown in Figure~\ref{fig:task}, debugging generally involves localizing errors, analyzing their causes, consulting schema definitions, applying targeted modifications, and re-running lint checks to verify whether requirements are satisfied—usually repeating this loop multiple times. Unfortunately, LLMs struggle with this iterative correction process. They frequently fall into anti-patterns such as repeating identical actions without meaningful follow-up, which leads to wasted effort when an initial correction fails \citep{bouzenia2025understandingsoftwareengineeringagents, laban2025llmslostmultiturnconversation}.

{\ul To bridge this gap, we propose moving beyond Text-to-SQL generation and shifting the focus to a model’s ability to iteratively debug and self-correct.} We introduce \ourbench, a benchmark for evaluating LLMs on enterprise-scale SQL debugging. Our construction pipeline uses an automated reverse-engineering framework to synthesize realistic, reproducible tasks. This approach minimizes human effort while ensuring high-quality benchmark generation, also providing a foundation for synthetic training data. Furthermore, we design an execution-free evaluation framework tailored to enterprise SQL scenarios. \ourbench offers a practical reference point for selecting SQL-focused LLMs in industry. The benchmark comprises $469$ \ourbenchsyn tasks (syntax errors with explicit error messages) and $516$ \ourbenchsem tasks (semantic errors in which the SQL output does not match the user’s requirement). SQL programs in our benchmark are highly complex, averaging over $140$ lines ($>420$ tokens), with ASTs of width $>11$ and depth $>8.7$, and incorporating over $15$ functions per script.

Our evaluation on \ourbench indicates significant room for improvement in deploying LLMs within SQL-SWE workflows. Extensive experiments show that even state-of-the-art LLMs struggle: Claude-4-Sonnet achieves only $36.46\%$ success on \ourbenchsyn and $33.17\%$ on \ourbenchsem, while most models fail to reach $20\%$. These results underscore the difficulty of enterprise SQL debugging and highlight substantial room for improvement. To address this gap, we systematically explore four potential solution strategies and conduct comprehensive experiments to assess their effectiveness. Our results not only illuminate the challenges LLMs face in SQL debugging but also offer insights into strategies to improve performance. Moreover, \ourbench exhibits a strong correlation with real-world debugging outcomes, establishing it as a reliable benchmark for aligning models with industrial applications. In summary, this work makes the following contributions:
\begin{itemize}[
    leftmargin=0.4cm,  
    itemindent=0.25cm,  
    labelsep=0.2cm,    
    topsep=0pt,        
    parsep=0pt         
    ]
\item We propose an automated reverse-engineering workflow for constructing high-quality SQL debugging benchmarks, which can also be adapted to synthesize realistic training data.
\item We present \ourbench, a large-scale benchmark comprising $469$ syntax and $516$ semantic tasks, designed to capture the complexity, diversity, and practicality of enterprise SQL development.
\item We conduct a comprehensive evaluation of nearly $30$ open-source and proprietary LLMs, showing that even the state-of-the-art LLMs face substantial challenges.
\item We introduce three SFT and an agent method as baselines, offering a novel and efficient pathway for further studies.
\end{itemize}

\paragraph{Conflict of Interest Disclosure}
Some of the authors are employees of ByteDance. The evaluated models include systems developed by a variety of organizations, including organizations affiliated with the authors' employer. All models were used solely for research and evaluation purposes under their applicable terms of use. The analyses and conclusions presented in this paper are independent academic findings and do not necessarily represent the views of any model developer or provider.
\section{Preliminary}
\subsection{Task Definition}
SQL debugging is a fundamental but underexplored problem in data development. Existing Text-to-SQL research primarily focuses on translating natural language to SQL queries, but real-world scenarios often involve correcting issues in SQL scripts. The goal of \task is to automatically repair buggy SQL scripts while preserving the user’s intent. This task begins with a buggy SQL query ($b$), accompanied by auxiliary context $\mathcal{C}$ (e.g., error messages or natural-language intent descriptions) and the database schema ($\sigma$). The objective is to generate a corrected SQL ($\hat q$):
\begin{equation}
\hat q = f_\theta(\mathcal{C}, \sigma, b)
\end{equation}
where $\hat q$ is syntax correct and faithful to the intent encoded in $(\mathcal{C}, b, \sigma)$.

We categorize bugs into two primary types: 
\textbf{(I) Syntactic errors.} $b$ is non-executable. Here, $\mathcal{C}$ is the error message $\mathcal{E}$, and the goal is to produce an executable repair while preserving its intended semantics.
\textbf{(II) Semantic errors.} $b$ executes successfully but fails to meet the user’s requirements. In this case, $\mathcal{C}$ is a natural language specification $\mathcal{R}$, and the task is to modify $\hat q$ to satisfy $\mathcal{R}$. By covering both types, \ourbench unifies execution repair with intent comprehension, offering a challenging and realistic benchmark for SQL debugging.

\subsection{Challenges}
Despite its practical importance, \task introduces several unique challenges that are not sufficiently addressed in existing SWE research.

\paragraph{Challenge 1: Lack of Enterprise-level SQL Scripts.}
Industrial SQL workloads, such as ETL workflows and scheduled analytical jobs, are typically \textit{long}, \textit{complex}, and \textit{schema-heavy}. Scripts can span hundreds of lines, involve deeply nested subqueries and multi-way joins. and reference dozens of tables and columns. This level of intricacy significantly amplifies the challenge for LLMs. In contrast, most existing Text-to-SQL \citep{li2024bird} and SQL-debugging \citep{bird-critic} benchmarks focus on short, relatively simple queries that are far removed from the scale and complexity of enterprise environments. Unfortunately, such industrial-grade SQL scripts are rarely available in the open-source community, resulting in a pronounced mismatch between academic benchmarks and real-world needs.
\begin{tcolorbox}[
    title={\textcolor{black}{\footnotesize \textbf{Contribution 1:}}},
    colframe=bytedsa!30,
    colback=bytedsa!5,
    boxsep=4pt,                    
    left=4pt, right=4pt,           
    top=2pt, bottom=2pt,           
    arc=2pt,              
    before skip=6pt, after skip=2pt 
    ]                      
\footnotesize
To address this gap, we introduce a large-scale, enterprise-level benchmark that captures the complexity of real-world ETL and analytical workloads (Section \ref{sec: gt sql creation}).
\end{tcolorbox}

\paragraph{Challenge 2:  Lack of a Comprehensive Bug Taxonomy.}
SQL bugs are heterogeneous: some manifest as execution failures (syntax errors), while others silently yield incorrect results (semantic errors). Although recent benchmarks such as BIRD-Critic \citep{bird-critic} have advanced debugging evaluation, they lack a systematic taxonomy of SQL-specific bug types and their prevalence. Without such categorization, it is difficult to understand where models struggle most and how to target improvements effectively. A comprehensive analysis of SQL bug categories is therefore crucial, not only for benchmarking but also for guiding the design of future bug-fixing models.
\begin{tcolorbox}[
    title={\textcolor{black}{\footnotesize \textbf{Contribution 2:}}},
    colframe=bytedsa!30,
    colback=bytedsa!5,
    boxsep=4pt,                    
    left=4pt, right=4pt,           
    top=2pt, bottom=2pt,           
    arc=2pt,              
    before skip=6pt, after skip=2pt 
    ]
\footnotesize
We develop a hierarchical taxonomy of SQL bug types derived from an extensive analysis of real-world errors. This provides a structured framework for fine-grained evaluation (Section \ref{sec: sql bug taxonomy}).
\end{tcolorbox}

\paragraph{Challenge 3: Lack of Reliable and Comprehensive \task Benchmark.}
High-quality benchmarks for SQL Debugging are scarce. Manually curated datasets are costly to produce and prone to evaluation leakage if models memorize solutions from public templates or repositories. Existing resources often lack diversity, realistic bug patterns, and coverage of enterprise-scale scripts, limiting their usefulness for robust model evaluation. Building a reliable, large-scale benchmark that is both comprehensive and faithful to real-world workflows is therefore a significant challenge.
\begin{tcolorbox}[
    title={\textcolor{black}{\footnotesize \textbf{Contribution 3:}}},
    colframe=bytedsa!30,
    colback=bytedsa!5,
    boxsep=4pt,                    
    left=4pt, right=4pt,           
    top=2pt, bottom=2pt,           
    arc=2pt,              
    before skip=6pt, after skip=2pt 
    ]
\footnotesize
We introduce an automated pipeline for synthesizing and validating SQL bug-fixing examples, ensuring scalability, diversity, and resistance to data leakage (Section \ref{sec: bug injection}).
\end{tcolorbox}

\section{\ourbench Construction}
\label{sec:benchmark}

\begin{figure*}[th]
    \centering
    \includegraphics[width=\linewidth]{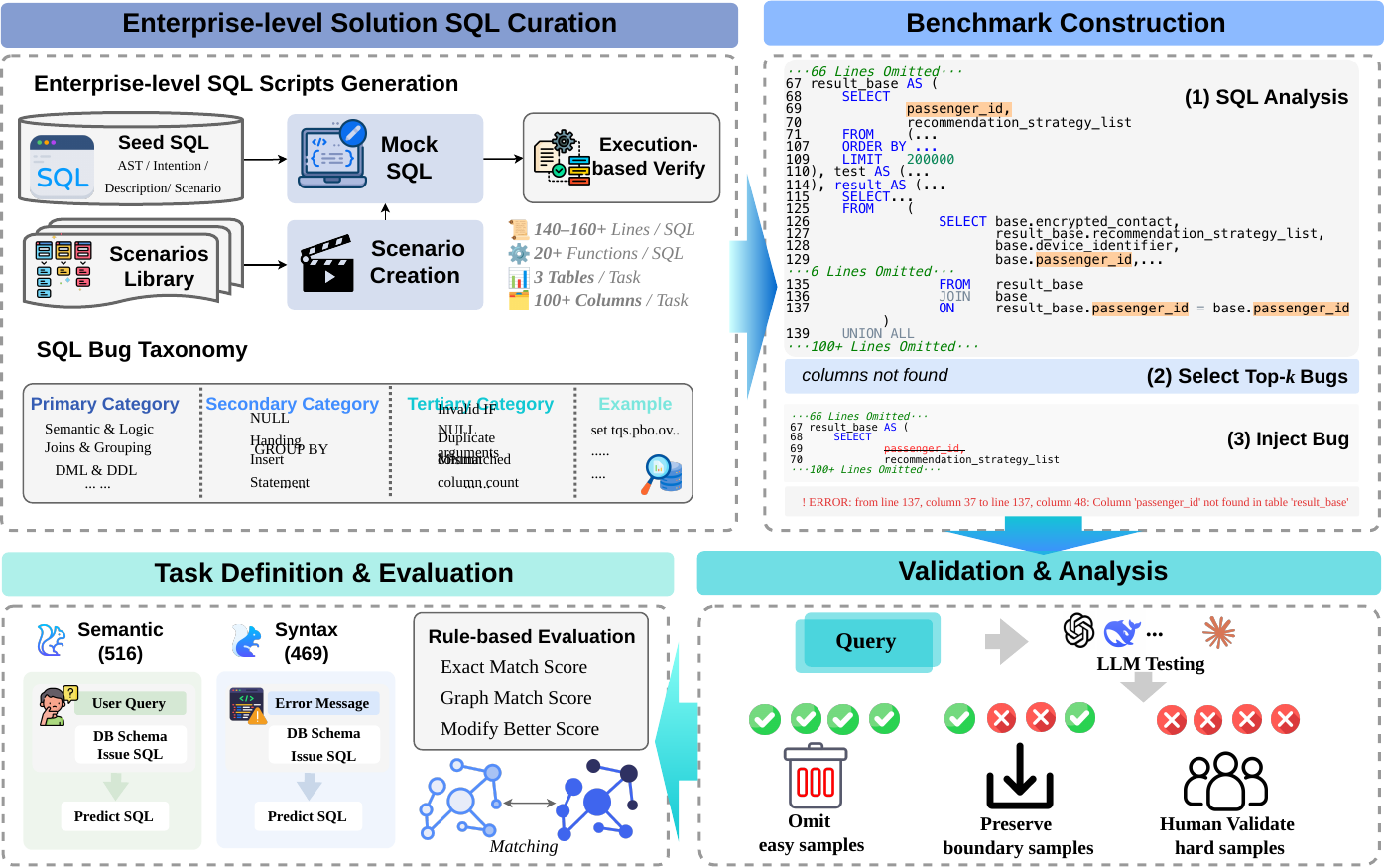}
    \caption{Overview of the \ourbench construction and evaluation pipeline. Benchmark construction consists of 4 main stages: (1) \textbf{Enterprise-level SQL Script Generation}, (2) \textbf{SQL bug taxonomy Design}, (3) \textbf{Issue SQL Construction via reverse engineering}, and (4) \textbf{Validation and Analysis}. This pipeline ensures diversity, realism, and rigorous evaluation of the SQL Debugging task.}
    \label{fig:overview}
\end{figure*}

Figure \ref{fig:overview} shows the automated benchmark construction pipeline. It comprises four stages: (1) enterprise-level SQL script generation (Section~\ref{sec: gt sql creation}), (2) SQL bug taxonomy design (Section~\ref{sec: sql bug taxonomy}), (3) issue SQL construction via reverse engineering (Section~\ref{sec: bug injection}), and (4) validation and analysis (Section~\ref{sec: validation}). 
Section~\ref{sec: evaluation metrics} further introduces an efficient execution-free evaluation methodology. 
Section~\ref{sec: evaluation metrics} presents an execution-free evaluation methodology. 
Examples and prompts are detailed in Appendix~\ref{sec:example} and ~\ref{sec:prompts}.

\subsection{Enterprise-level SQL Scripts Generation}
\label{sec: gt sql creation}
Because enterprise SQL scripts are proprietary and rarely accessible, we synthesize realistic, high-quality enterprise SQL.

\paragraph{Seed Enterprise SQL Curation.}  

We curate high-quality SQL scripts $q$ along with corresponding table definitions $\sigma$ from real-world enterprise applications. To ensure that queries are non-trivial and representative of practical workloads, we filter scripts that fall below a complexity threshold $\tau$. Complexity is quantified via a composite metric:
\begin{equation}
\mathcal{C}(q) = \alpha \bigl( D_{\text{AST}}(q) + W_{\text{AST}}(q) \bigr) + \beta L(q)
\end{equation}
where $D_{\text{AST}}$, $W_{\text{AST}}$, and $L(q)$ denote AST depth, AST width, and code length, respectively. 

For each retained SQL script, we utilize an LLM to abstract its business domain ($d$), intention ($I$), and descriptive scenario ($S$). All scenarios are aggregated into a Scenarios Library, denoted as $\mathcal{D}_{\text{domain}} = \{d\}$. The resulting seed dataset is then defined as:
\begin{equation}
\mathcal{D}_{\text{seed}} = \{ (q_i, \sigma_i, d_i, I_i, S_i, \text{AST}(q_i)) | q_i \in \mathcal{Q}_s, \mathcal{C}(q_i) > \tau \},
\end{equation}
where $\mathcal{Q}_s$ denotes the candidate SQL pool.

The final seed corpus contains $1{,}000+$ SQL scripts spanning $26$ business scenarios, averaging over $120$ lines with AST depth $>8$ and width $>12$. Each script is rigorously validated to be bug-free, resulting in a corpus that accurately captures both the structural complexity and semantic diversity of enterprise SQL.

\paragraph{Solution SQL Synthesis.}  

To expand coverage across domains and code structures, we synthesize new SQL scripts using the seed corpus and the Scenarios Library:

1. \emph{Seed Sampling.}  
   Select $(q_i, \sigma_i, d_i, I_i, S_i, \text{AST}(q_i)) \in \mathcal{D}_{\text{seed}}$ and a target domain $d_t \in \mathcal{D}_{\text{domain}}$.  

2. \emph{Scenario Creation.}  
   Conditioned on $d_t$, the LLM generates a new scenario description $\mathcal{S}_t$ together with schema definitions $\sigma_t$, following the structure of the seed corpus.  

3. \emph{SQL Synthesis.}  
   Given $(I_i, S_i, \text{AST}(q_i), \mathcal{S}_t, \sigma_t)$, the LLM generates a new SQL script $q_t$ that preserves the complexity of the seed SQL scripts while adapting to the new schema and scenario. This ensures that synthesized queries remain realistic, non-trivial, and representative of enterprise workloads.  

4. \emph{Execution-based Validation.}  
   To ensure the correction, each candidate $q_t$ is validated via execution. Specifically, $\sigma_t$ is instantiated to construct a fake test database, $q_t$ is executed, and only queries that successfully execute are retained:  
\begin{equation}
     \mathcal{Q}_{\text{gt}} = \bigl\{ (q_t, \sigma_t) \;\big|\; \text{exec}(q_t,\sigma_t) == \text{passed} \bigr\}
\end{equation}
This synthesis pipeline ensures that the final SQL dataset exhibits (i) enterprise-grade complexity, (ii) broad domain coverage via controlled scenario transfer, and (iii) guaranteed execution correctness.

\subsection{SQL Bug Taxonomy}
\label{sec: sql bug taxonomy}

We construct an SQL bug taxonomy by manually annotating $268$ erroneous SQL scripts collected from real-world production logs. Each bug is classified according to a three-level hierarchical error type: 
(i) \emph{macro categories} (e.g., DML, DDL, semantic, and logic),  
(ii) \emph{construct-specific subcategories} (e.g., \texttt{INSERT} statements), and  
(iii) \emph{atomic faults} (e.g., mismatched column counts).  
This taxonomy organizes common failure patterns and forms a bug library of realistic error templates. The library underpins our controlled bug-injection process (Section~\ref{sec: bug injection}), ensuring that \ourbench captures authentic SQL error modes. Table~\ref {tab:syntax} and~\ref {tab:semantic} report the distribution of bug types.
 
\subsection{Issue SQL Construction}
\label{sec: bug injection}
We construct issue SQL queries through reverse engineering, transforming correct SQL scripts into buggy versions. The process is guided by three principles: structural awareness, taxonomy-guided selection, and minimal-change injection, ensuring that the generated bugs are both realistic and diagnostically useful.

\textbf{Step 1: Structural Profiling and Taxonomy-Guided Selection}
For each ground-truth SQL $q_{\text{gt}}$, we first analyze its structural and semantic profile, including the AST, function patterns, and clause usage. Based on this profile, we then select the top-$k$ candidate bug types from our hierarchical SQL bug taxonomy. This approach ensures that the injected errors are well-suited to the given SQL while providing broad coverage of real-world error scenarios.

\textbf{Step 2: Minimal Change-Based Bug Injection.}
Each injected bug represents the smallest possible modification that induces the targeted error type. This principle preserves maximal similarity between the buggy SQL $b$ and its reference $q_{\text{gt}}$, isolating the error signal and reducing confounding factors. As a result, evaluating whether a model can localize and repair the fault becomes both precise and interpretable.

\subsection{Validation and Analysis}
\label{sec: validation}
We validate \ourbench via a model-driven \textit{attack–defense} process. The goal is to filter out trivial cases that most models can easily solve, while retaining challenging but solvable instances that better reflect real-world debugging.

\paragraph{Automated Verification.}
We first attack the benchmark by evaluating each generated instance with a diverse set of advanced LLMs (including Qwen3-Coder-32B\citep{yang2025qwen3}, GPT-5\citep{gpt-5}, DeepSeek-V3.1\citep{deepseek-v3.1}, Claude-4-sonnet\citep{claude}, and others). Instances fall into three categories:
(i) If the majority of models succeed, the instance is deemed too easy and discarded;
(ii) If only a few models succeed, the instance is considered an edge case and retained;
(iii) If none of the models succeed, the instance is flagged for manual review.
This adversarial filtering ensures that the benchmark emphasizes cases where current models diverge, thereby sharpening its discriminatory power.

\paragraph{Human Verification.}
Instances flagged as potentially unsolvable are subjected to manual inspection by three expert annotators with extensive SQL experience. Following a cross-validation protocol, annotators assess whether the task is logically inferable from the provided context and whether multiple valid solutions exist. Instances that fail to meet these criteria are removed. For cases where multiple correct answers are possible, annotators supplement the benchmark with all valid alternative solutions.

Through this \textit{attack–defense} protocol, \ourbench removes trivial cases, yielding a challenging yet solvable testbed.

\subsection{Evaluation Metrics}
\label{sec: evaluation metrics}
The prevailing metrics for \task are Exact Match (EM) and Execution Accuracy. However, EM is notoriously strict, failing to credit semantically equivalent queries with divergent syntax. Execution Accuracy, while more forgiving, introduces false positives when test databases lack the necessary content to reveal logical errors \citep{zhan-etal-2025-towards}.  Direct execution in production also poses practical barriers, being computationally expensive and raising data privacy concerns. To overcome these challenges, we introduce an execution-free evaluation framework based on three metrics (Detailed definitions and formulas are provided in Appendix~\ref{app: evaluation metrics}.):

 (1) \textbf{Exact Match Score (EM)}: This metric assesses strict syntactic correctness by checking for string-level identity between the predicted and reference SQL queries, thereby serving as a baseline for syntactic alignment.

 (2) \textbf{Graph Match Score (GM)}: This metric evaluates structural and functional equivalence by comparing the optimized abstract syntax tree of the predicted and reference queries, thereby capturing semantic correctness where EM fails.
 
 (3) \textbf{Modify Better Score (MB)}: This metric gauges iterative improvement capability by comparing the edit distances from the predicted SQL and the original SQL to the reference, thereby measuring how much closer the refinement is to the target.

\section{Benchmark Statistics}
\label{sec:benchmark statistic}

\begin{figure*}[tb]
    \centering
    \begin{minipage}{0.26\textwidth}
        \centering
        \includegraphics[width=\textwidth]{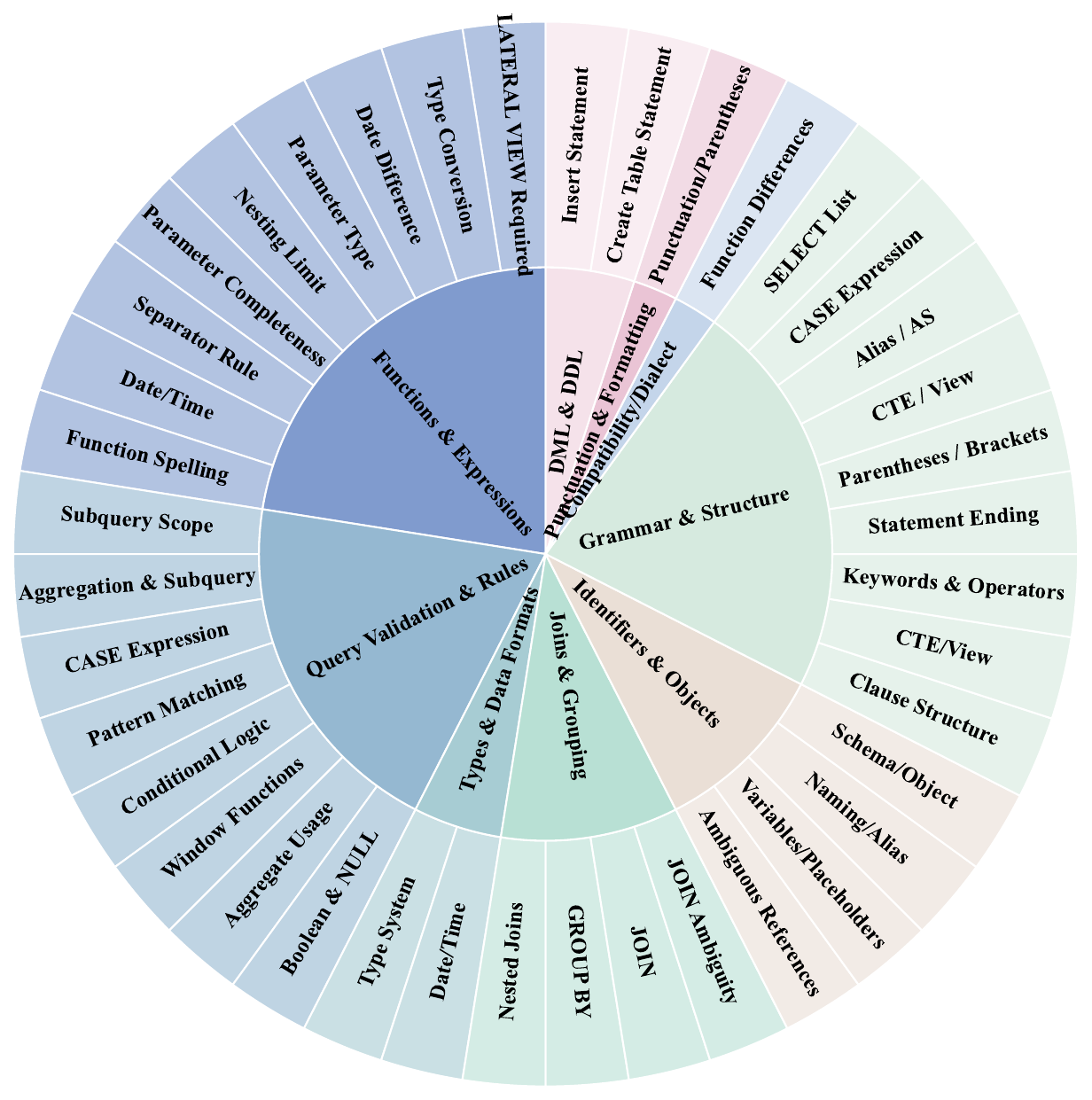}
        \caption*{(a)}
    \end{minipage}
    \hfill
    \begin{minipage}{0.26\textwidth}
        \centering
        \includegraphics[width=\textwidth]{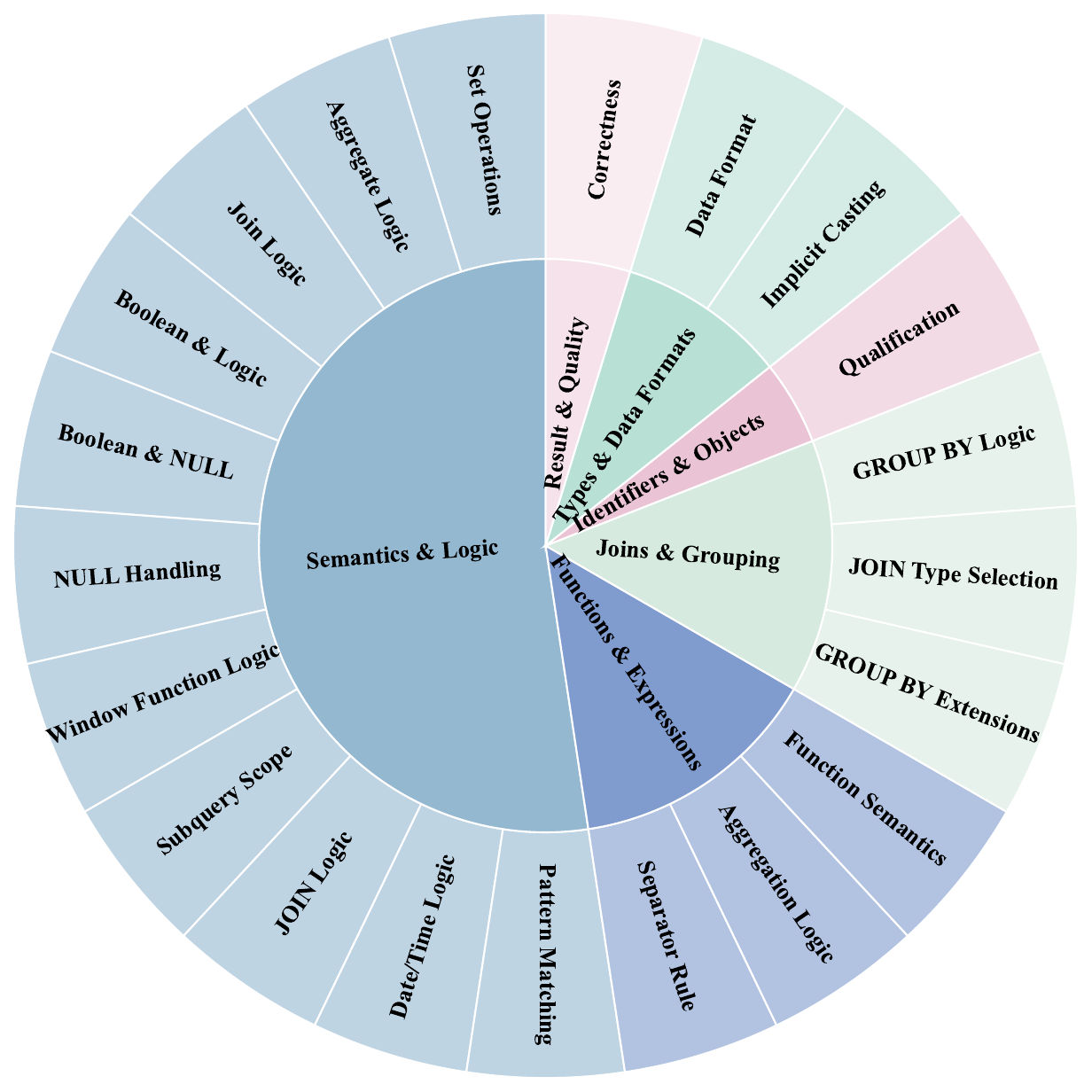}
        \caption*{(b)}
    \end{minipage}
    \hfill
    \begin{minipage}{0.3\textwidth}
        \centering
        \includegraphics[width=\textwidth]{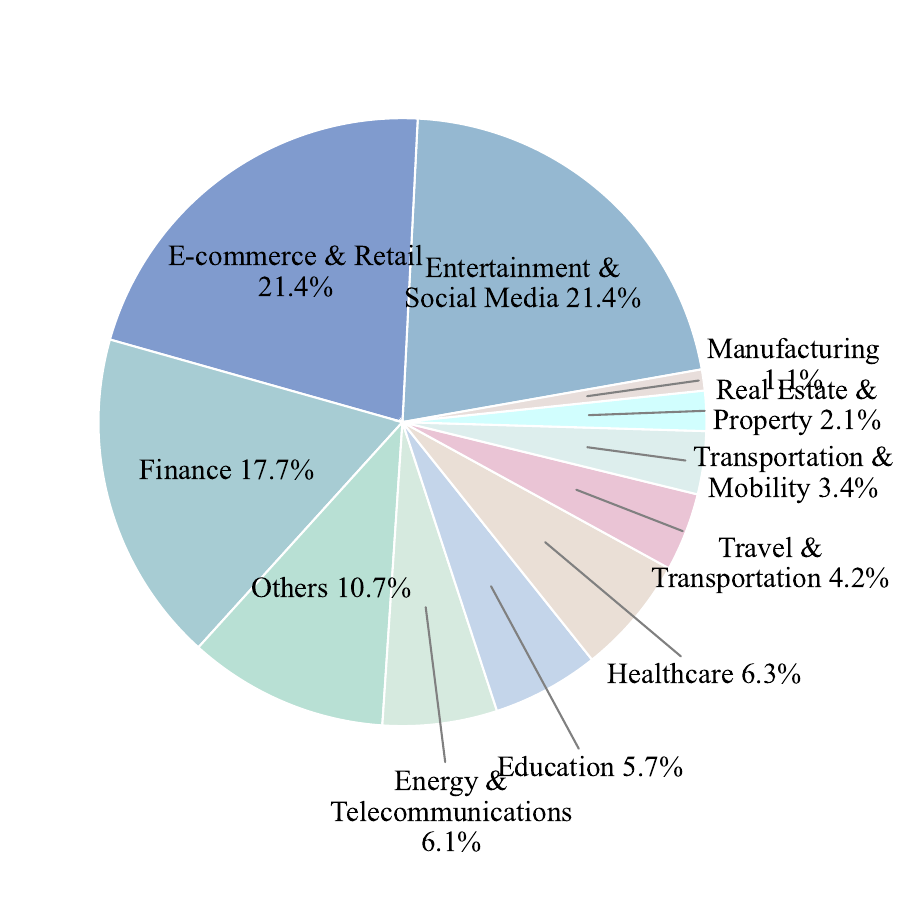}
    \end{minipage}
    \caption{Statistics of errors and domain distribution in \ourbench. (a) Two-level error types in \ourbenchsyn, highlighting the distribution of syntax errors. 
(b) Two-level error types in \ourbenchsem, showing the distribution of semantic errors. 
(c) Distribution of SQL code across different business domains.}
    \label{fig:statistic}
\end{figure*}

We present a statistical analysis of \ourbench, comparing it with existing SQL datasets in Table~\ref{tab:statistic} and Figure~\ref{fig:statistic}. The benchmark emphasizes \textit{complexity} and \textit{realism}, closely reflecting real-world industrial SQL challenges in script structure, error taxonomy, and task diversity.

\begin{table*}[htb]
\footnotesize
\centering
\setstretch{1.2}
\caption{Statistical comparison of \ourbench with representative text-to-SQL and SQL debugging benchmarks. The table evaluates benchmarks on scale (\# examples), script length (avg. tokens and lines), and structural complexity (avg. function count, AST depth, and width).}
\label{tab:statistic}
\begin{adjustbox}{width=\textwidth}
\begin{tabular}{@{\hskip 0pt} l c c c c c c c @{\hskip 0pt}}
\toprule[1.3pt]
\multicolumn{1}{c}{} & & & \multicolumn{2}{c}{\textbf{Length of SQL}} & \multicolumn{3}{c}{\textbf{Complex of SQL}} \\ 
\cmidrule(l){4-5} \cmidrule(l){6-8} 
\multicolumn{1}{c}{\multirow{-2}{*}{\textbf{Benchmark}}} &
\multirow{-2}{*}{\textbf{Type}} &
\multirow{-2}{*}{\textbf{\makecell[c]{\# Test \\ Examples}}} &
\textbf{\makecell[c]{\# Tok. \\ /SQL}} &
\textbf{\makecell[c]{\# Line. \\ /SQL}} &
\textbf{\makecell[c]{\# Func. \\ /SQL}} &
\textbf{\makecell[c]{\# AST Depth \\ /SQL}} &
\textbf{\makecell[c]{\# AST Width \\ /SQL}} \\ \midrule
Spider 1.0  \citep{yu2018spider}           & Text-to-SQL   & 2,147 & 18.50 & — & — & — & — \\
Spider 2.0-snow \citep{lei2025spider}       & Text-to-SQL   & 121  & 154.63 & 56.12 & 14.90 & \textbf{11.95} & 9.66 \\
Spider 2.0-lite \citep{lei2025spider}       & Text-to-SQL   & 256  & 131.79 & 49.84 & 13.65 & \textbf{11.97} & 10.05 \\
BIRD \citep{li2024bird}                    & Text-to-SQL   & 1,789 & 30.90 & — & — & — & — \\ \midrule
BIRD-Critic-open \citep{bird-critic}       & SQL debugging & 600  & 49.18 & 9.73 & 4.30 & 8.03 & 6.01 \\
BIRD-Critic-postgresql \citep{bird-critic} & SQL debugging & 530  & 51.44 & 6.92 & 4.78 & 8.25 & 6.34 \\
BIRD-Critic-flash \citep{bird-critic}      & SQL debugging & 200  & 34.53 & 2.84 & 4.06 & 7.85 & 5.20 \\ \midrule
\rowcolor[HTML]{ECF4FF} 
Squirrel-Syntax                            & SQL debugging & 469  & \textbf{496.90} & \textbf{163.69} & \textbf{21.62} & 8.93 & \textbf{11.69} \\
\rowcolor[HTML]{ECF4FF} 
Squirrel-Semantic                          & SQL debugging & 516  & \textbf{425.93} & \textbf{141.58} & \textbf{17.34} & 8.75 & \textbf{11.12} \\ \bottomrule[1.3pt]
\end{tabular}
\end{adjustbox}
\end{table*}

\paragraph{Complexity of SQL Scripts.}
The SQL scripts in \ourbench are not only longer but also structurally more complex, presenting challenges that better mirror real-world enterprise systems. With an average length of $140-160$ lines and over $420$ tokens, our scripts are an order of magnitude larger than those in BIRD-Critic (which average under $10$ lines). This scale directly implies a higher probability of errors and a greater need for models to maintain long-range context and dependency understanding. Additionally, the high number of functions per script ($17.34$ in \ourbenchsem, $21.62$ in \ourbenchsyn) necessitates reasoning across multiple subqueries and nested expressions—a capability that many existing sequence-to-sequence models lack. This scale and functional richness underscore the increased complexity and practical difficulty of the debugging tasks in our benchmark.

\paragraph{Hierarchical Error Taxonomy.}
Figures~\ref{fig:statistic}(a) and (b) show the two-level error taxonomy for \ourbenchsyn and \ourbenchsem. Detailed error type statistics are in Appendix~\ref{app:SQL Bug Taxonomy}. \ourbench covers a broad spectrum of common syntax and semantic errors, enabling fine-grained evaluation of model capabilities. Syntax errors include issues related to grammar, structure, and dialect, while semantic errors encompass type mismatches, aggregation errors, and logical inconsistencies. This hierarchical classification allows for detailed insight into model performance across error types, supporting a more rigorous assessment of debugging ability.

\paragraph{Diversity of Task Scenarios.}
As shown in Figure~\ref{fig:statistic} (c), the domains in \ourbench span finance, e-commerce, healthcare, and more than ten additional areas, ensuring that models are evaluated against a broad range of business logic and contextual dependencies. For example, a program from the financial domain may involve complex window functions for time-series analysis, whereas an e-commerce program might require reasoning over multi-table joins across user and product schemas. This diversity tests a model’s ability to generalize beyond simplistic syntactic patterns and demands domain-aware reasoning. Consequently, performance on \ourbench provides a stronger indicator of a model’s practicality and readiness for deployment in heterogeneous real-world environments.

\section{Experiments}
\begin{table*}[thb]
\setstretch{1}
\footnotesize
\centering
\caption{Evaluation results of LLMs on \ourbenchsyn and \ourbenchsem. For each section, the best performance is highlighted in \textbf{bold}, and the second-best is {\ul underlined}. EM, GM, and MB denote exact match score, graph match score, and modify-better score, respectively.}
\label{tab:main_results}
\begin{adjustbox}{width=\linewidth}
\begin{tabular}{lccc S S S S S S}
\toprule[1.2pt]
\multicolumn{1}{c|}{} &
  \multicolumn{1}{c|}{} &
   &
  \multicolumn{1}{c|}{} &
  \multicolumn{3}{c|}{\textbf{Squirrel-Syntax}} &
  \multicolumn{3}{c}{\textbf{Squirrel-Semantic}} \\ \cmidrule{5-10} 
\multicolumn{1}{c|}{\multirow{-2}{*}{\textbf{Model}}} &
  \multicolumn{1}{c|}{\multirow{-2}{*}{\textbf{Size}}} &
  \multirow{-2}{*}{\textbf{Reasoning}} &
  \multicolumn{1}{c|}{\multirow{-2}{*}{\textbf{MoE}}} &
  \textbf{EM} &
  \textbf{GM} &
  \multicolumn{1}{c|}{\textbf{MB}} &
  \textbf{EM} &
  \textbf{GM} &
  \textbf{MB} \\ \midrule
\multicolumn{10}{l}{\cellcolor[HTML]{CDDBFC}{\color[HTML]{1F2329} \textit{\textbf{Open Source}}}} \\
\multicolumn{1}{l|}{Qwen-2.5-Instruct} &
  \multicolumn{1}{c|}{7B} &
   &
  \multicolumn{1}{l|}{} &
  2.13 &
  8.53 &
  \multicolumn{1}{c|}{33.05} &
  1.94 &
  5.62 &
  14.15 \\
\multicolumn{1}{l|}{Qwen-2.5-Coder} &
  \multicolumn{1}{c|}{7B} &
   &
  \multicolumn{1}{c|}{} &
  3.20 &
  8.96 &
  \multicolumn{1}{c|}{37.53} &
  4.84 &
  7.75 &
  18.99 \\
\multicolumn{1}{l|}{Qwen-2.5-Coder} &
  \multicolumn{1}{c|}{32B} &
   &
  \multicolumn{1}{c|}{} &
  12.79 &
  20.26 &
  \multicolumn{1}{c|}{52.88} &
  \textbf{17.44} &
  \textbf{23.45} &
  \textbf{34.69} \\
\multicolumn{1}{l|}{Qwen-3-Instruct} &
  \multicolumn{1}{c|}{235B} &
   &
  \multicolumn{1}{c|}{$\checkmark$} &
  9.38 &
  20.47 &
  \multicolumn{1}{c|}{61.19} &
  10.27 &
  15.50 &
  27.57 \\
\multicolumn{1}{l|}{Qwen-3-Coder-Instruct} &
  \multicolumn{1}{c|}{30B} &
   &
  \multicolumn{1}{c|}{$\checkmark$} &
  5.54 &
  20.90 &
  \multicolumn{1}{c|}{44.14} &
  6.40 &
  15.12 &
  24.42 \\
\multicolumn{1}{l|}{Qwen-3-Coder-Instruct} &
  \multicolumn{1}{c|}{480B} &
   &
  \multicolumn{1}{c|}{$\checkmark$} &
  14.93 &
  23.88 &
  \multicolumn{1}{c|}{61.62} &
  17.05 &
  19.96 &
  31.84 \\
\multicolumn{1}{l|}{QwQ} &
  \multicolumn{1}{c|}{32B} &
  $\checkmark$ &
  \multicolumn{1}{c|}{} &
  8.76 &
  20.51 &
  \multicolumn{1}{c|}{41.45} &
   10.47&
   15.31&
   20.16\\
\multicolumn{1}{l|}{Seed-Coder-Instruct} &
  \multicolumn{1}{c|}{8B} &
   &
  \multicolumn{1}{c|}{} &
   8.53&
   14.93&
  \multicolumn{1}{c|}{42.43} &
   8.72&
   14.15&
   24.61\\
\multicolumn{1}{l|}{OmniSQL} &
  \multicolumn{1}{c|}{32B} &
   &
  \multicolumn{1}{c|}{} &
  0.21 &
  6.40 &
  \multicolumn{1}{c|}{50.75} &
  0.39 &
  6.40 &
  21.17 \\
\multicolumn{1}{l|}{Deepseek-V3} &
  \multicolumn{1}{c|}{685B} &
   &
  \multicolumn{1}{c|}{$\checkmark$} &
  17.91 &
  \textbf{30.28} &
  \multicolumn{1}{c|}{60.34} &
  11.24 &
  21.32 &
  33.27 \\
\multicolumn{1}{l|}{Deepseek-V3.1} &
  \multicolumn{1}{c|}{685B} &
   &
  \multicolumn{1}{c|}{$\checkmark$} &
  17.91 &
  30.49 &
  \multicolumn{1}{c|}{\textbf{63.61}} &
  12.02 &
  14.73 &
  32.47 \\
\multicolumn{1}{l|}{Deepseek-R1} &
  \multicolumn{1}{c|}{671B} &
  $\checkmark$ &
  \multicolumn{1}{c|}{$\checkmark$} &
  \textbf{18.34} &
  21.98 &
  \multicolumn{1}{c|}{58.64} &
  15.89 &
  22.09 &
  30.14 \\ \midrule
\multicolumn{10}{l}{\cellcolor[HTML]{E1EAFF}\textit{\textbf{Closed Source}}} \\
\multicolumn{1}{l|}{Claude-4-Sonnet} &
  \multicolumn{1}{c|}{—} &
  $\checkmark$ &
  \multicolumn{1}{c|}{\textbf{}} &
  \textbf{23.88} &
  \textbf{36.46} &
  \multicolumn{1}{c|}{\textbf{68.02}} &
  \textbf{31.78} &
  \textbf{32.17} &
  \textbf{43.69} \\
\multicolumn{1}{l|}{GPT-4o-mini-2024-07-18} &
  \multicolumn{1}{c|}{—} &
   &
  \multicolumn{1}{c|}{} &
  1.71 &
  4.69 &
  \multicolumn{1}{c|}{13.01} &
  5.62 &
  6.40 &
  8.74 \\
\multicolumn{1}{l|}{GPT-4o-2024-11-20} &
  \multicolumn{1}{c|}{—} &
   &
  \multicolumn{1}{c|}{} &
  2.14 &
  4.69 &
  \multicolumn{1}{c|}{13.79} &
  2.91 &
  4.84 &
  6.86 \\
\multicolumn{1}{l|}{GPT-4.1} &
  \multicolumn{1}{c|}{—} &
   &
  \multicolumn{1}{c|}{} &
  6.40 &
  17.70 &
  \multicolumn{1}{c|}{61.25} &
  8.52 &
  17.05 &
  30.49 \\
\multicolumn{1}{l|}{GPT-5} &
  \multicolumn{1}{c|}{—} &
  $\checkmark$ &
  \multicolumn{1}{c|}{} &
  13.43 &
  18.55 &
  \multicolumn{1}{c|}{66.52} &
  16.28 &
  16.47 &
  29.90 \\
\multicolumn{1}{l|}{Gemini-2.5-Pro} &
  \multicolumn{1}{c|}{—} &
  $\checkmark$ &
  \multicolumn{1}{c|}{} &
  15.78 &
  21.54 &
  \multicolumn{1}{c|}{62.37} &
  14.15 &
  23.06 &
  34.37 \\
\multicolumn{1}{l|}{Kimi-K2} &
  \multicolumn{1}{c|}{—} &
  $\checkmark$ &
  \multicolumn{1}{c|}{$\checkmark$} &
  14.07 &
  27.72 &
  \multicolumn{1}{c|}{61.83} &
  15.70 &
  20.93 &
  31.84 \\
\multicolumn{1}{l|}{O1-preview} &
  \multicolumn{1}{c|}{—} &
  $\checkmark$ &
  \multicolumn{1}{c|}{} &
  8.32 &
  21.11 &
  \multicolumn{1}{c|}{46.27} &
   8.14&
   11.43&
   14.43\\
\multicolumn{1}{l|}{O3-mini} &
  \multicolumn{1}{c|}{—} &
  $\checkmark$ &
  \multicolumn{1}{c|}{} &
  3.84 &
  19.83 &
  \multicolumn{1}{c|}{63.54} &
  10.47 &
  28.68 &
  40.78 \\
\multicolumn{1}{l|}{Doubao-Seed-1.6} &
  \multicolumn{1}{c|}{230B} &
  $\checkmark$ &
  \multicolumn{1}{c|}{$\checkmark$} &
  19.19 &
  30.92 &
  \multicolumn{1}{c|}{64.39} &
  16.09 &
  20.93 &
  32.82 \\
\multicolumn{1}{l|}{Doubao-Seed-1.6-flash} &
  \multicolumn{1}{c|}{230B} &
  $\checkmark$ &
  \multicolumn{1}{c|}{$\checkmark$} &
  1.50 &
  3.63 &
  \multicolumn{1}{c|}{9.62} &
  1.55 &
  3.11 &
  6.42 \\
\multicolumn{1}{l|}{Doubao-Seed-1.6-thinking} &
  \multicolumn{1}{c|}{230B} &
  $\checkmark$ &
  \multicolumn{1}{c|}{$\checkmark$} &
  15.35 &
  23.24 &
  \multicolumn{1}{c|}{60.98} &
  16.67 &
  20.93 &
  30.87 \\ \midrule
\multicolumn{10}{l}{\cellcolor[HTML]{D9F3FD}\textit{\textbf{Comparison of different SFT method on Qwen-2.5-Coder}}} \\
\multicolumn{1}{l|}{+ SFT} &
  \multicolumn{1}{c|}{} &
  \mbox{\ul{}} &
  \multicolumn{1}{c|}{\mbox{\ul{}}} &
  \mbox{\ul{26.44}} &
  \mbox{\ul{30.70}} &
  \multicolumn{1}{c|}{\mbox{\ul{48.40}}} &
  \mbox{\ul{14.34}} &
  \mbox{\ul{15.70}} &
  \mbox{\ul{18.02}} \\
\multicolumn{1}{l|}{+ diff-SFT} &
  \multicolumn{1}{c|}{} &
   &
  \multicolumn{1}{c|}{} &
  22.17 &
  22.81 &
  \multicolumn{1}{c|}{34.33} &
  7.95 &
  9.30 &
  12.60 \\
\multicolumn{1}{l|}{+ DM-SFT} &
  \multicolumn{1}{c|}{\multirow{-3}{*}{7B}} &
  \textbf{} &
  \multicolumn{1}{c|}{\textbf{}} &
  \textbf{27.27} &
  \textbf{33.18} &
  \multicolumn{1}{c|}{\textbf{55.67}} &
  \textbf{15.12} &
  \textbf{18.99} &
  \textbf{24.81} \\ \bottomrule[1.2pt]
\end{tabular}
\end{adjustbox}
\end{table*}

Detailed experimental settings are in Appendix~\ref{app:experimental settings}; this section highlights key results.

\subsection{Main Results}
\label{sec: results}

\paragraph{Existing LLMs are far from being experts on enterprise \task.}  
As shown in Table~\ref{tab:main_results}, we evaluate a diverse set of LLMs on Squirrel, including the Qwen, DeepSeek, Claude, GPT, Gemini, and Doubao families. Claude-4-Sonnet achieves the best performance, with a peak success rate of {$36.46\%$} GM score on Squirrel-Syntax and {$32.17\%$} GM score on Squirrel-Semantic. 
Other closed-source LLMs perform even worse, with most failing to exceed $20\%$ GM. Among open-source models, DeepSeek-V3 achieves $30.28\%$ on \ourbenchsyn, and Qwen-2.5-Coder-32B attains $23.45\%$ on \ourbenchsem, demonstrating competitive performance relative to closed-source systems.

\paragraph{Code generation LLMs struggle with SQL debugging.}  
In previous studies, most code LLMs are heavily optimized for code generation, achieving strong performance on benchmarks such as SWE-Bench~\citep{swebench}, BIRD~\citep{li2024bird}, and Spider~\citep{yu2018spider}. For example, OmniSQL~\citep{li2025omnisql}, a Text-to-SQL–specialized model, achieves $87.6\%$ on Spider and $64.5\%$ on BIRD. However, its performance on \ourbenchsyn and \ourbenchsem drops sharply to only $6.4\%$ GM, underscoring the substantial gap between SQL generation and SQL debugging.

\paragraph{Reasoning-oriented LLMs (RLMs) exhibit stronger refinement abilities.}  
Comparing RLMs with non-RLMs, we find that RLMs consistently perform better across both open-source and closed-source families. Notably, most RLMs achieve MB scores above $50\%$, indicating that while their predictions often move closer to the correct solution, they rarely solve the task in a single attempt.

\paragraph{Squirrel-Semantic is more challenging than Squirrel-Syntax.}  
Across all evaluated models, performance on Squirrel-Semantic is consistently lower than on Squirrel-Syntax. This is because Squirrel-Syntax provides explicit error messages, which help models localize faulty positions, whereas Squirrel-Semantic requires reasoning about deeper semantic inconsistencies without surface-level cues.

\subsection{Can SFT Solve the \task?}

\begin{figure}[ht]
    \centering
    \includegraphics[width=\linewidth]{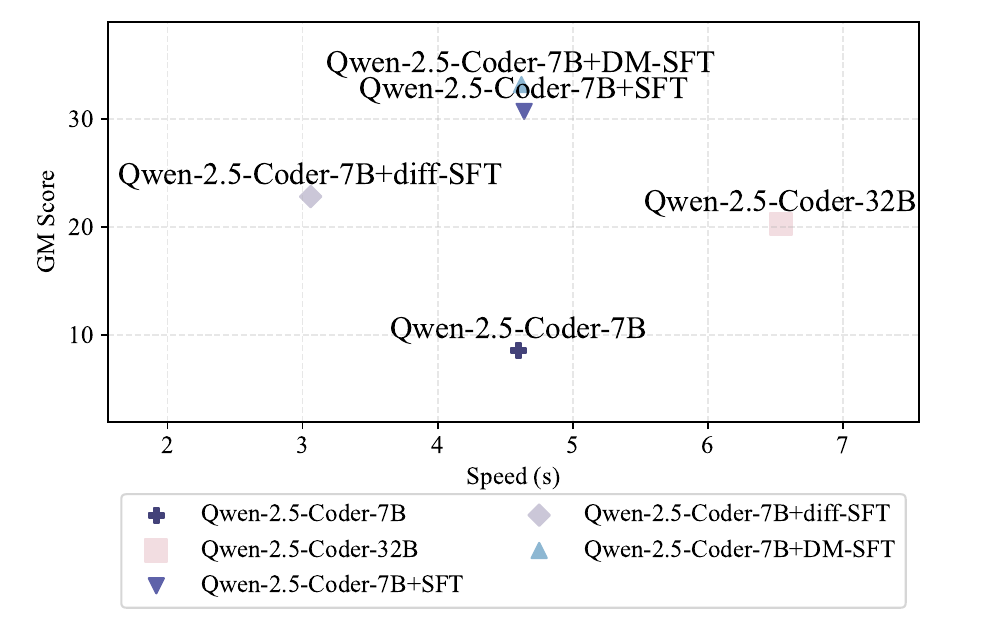}
    \caption{SFT baseline performance on \ourbenchsyn. The horizontal axis represents the average inference speed, and the vertical axis shows the GM score.}
    \label{fig:sft perfomance comparsion}
\end{figure}

As detailed in Appendix~\ref{app:sft baseline} and Figure~\ref{fig:sft methods}, we propose 3 representative SFT approaches as baselines: \textbf{(1) Vanilla SFT}, which directly fine-tunes the model on parallel SQL debugging pairs; 
\textbf{(2) DM-SFT}~\citep{duan2024pdcdmsftroad}, which dynamically masking the loss for unchanged tokens in responses; 
\textbf{(3) Diff-SFT}, which frames SFT as a search-and-replace task, focusing only on the modified code segments. 
Results in Table~\ref{tab:main_results} and Figure~\ref{fig:sft perfomance comparsion} shows: 

\textbf{(1)} Targeted in-domain SFT significantly improves SQL debugging performance. Specifically, Qwen-2.5-Coder-7B + SFT substantially outperforms the base Qwen-2.5-Coder-7B, achieving a $33.17\%$ gain in GM score on Squirrel-Syntax, and even surpasses Qwen-2.5-Coder-32B by $10.44\%$. 
\textbf{(2)} DM-SFT improves performance over vanilla SFT by masking the loss on non-diff tokens during training. This design forces the model to focus more on diff segments within pairs, thereby enhancing its effectiveness.
\textbf{(3)} Diff-SFT predicts only the diff segments instead of generating the full code, offering a substantial inference speed advantage and reducing generation hallucination. On our benchmark, it requires only half the time of other methods, which is particularly beneficial for longer code snippets in enterprise applications. However, due to a mismatch between the search-and-replace task and the pretraining/SFT objectives of the base model, its GM score is slightly lower. 
Overall, these three SFT strategies provide strong baselines for future research on \task. More analysis is available in Appendix~\ref{app:sft}.

\subsection{Can Agent Methods Solve the \task?}

\begin{figure}[ht]
    \centering
    \includegraphics[width=\linewidth]{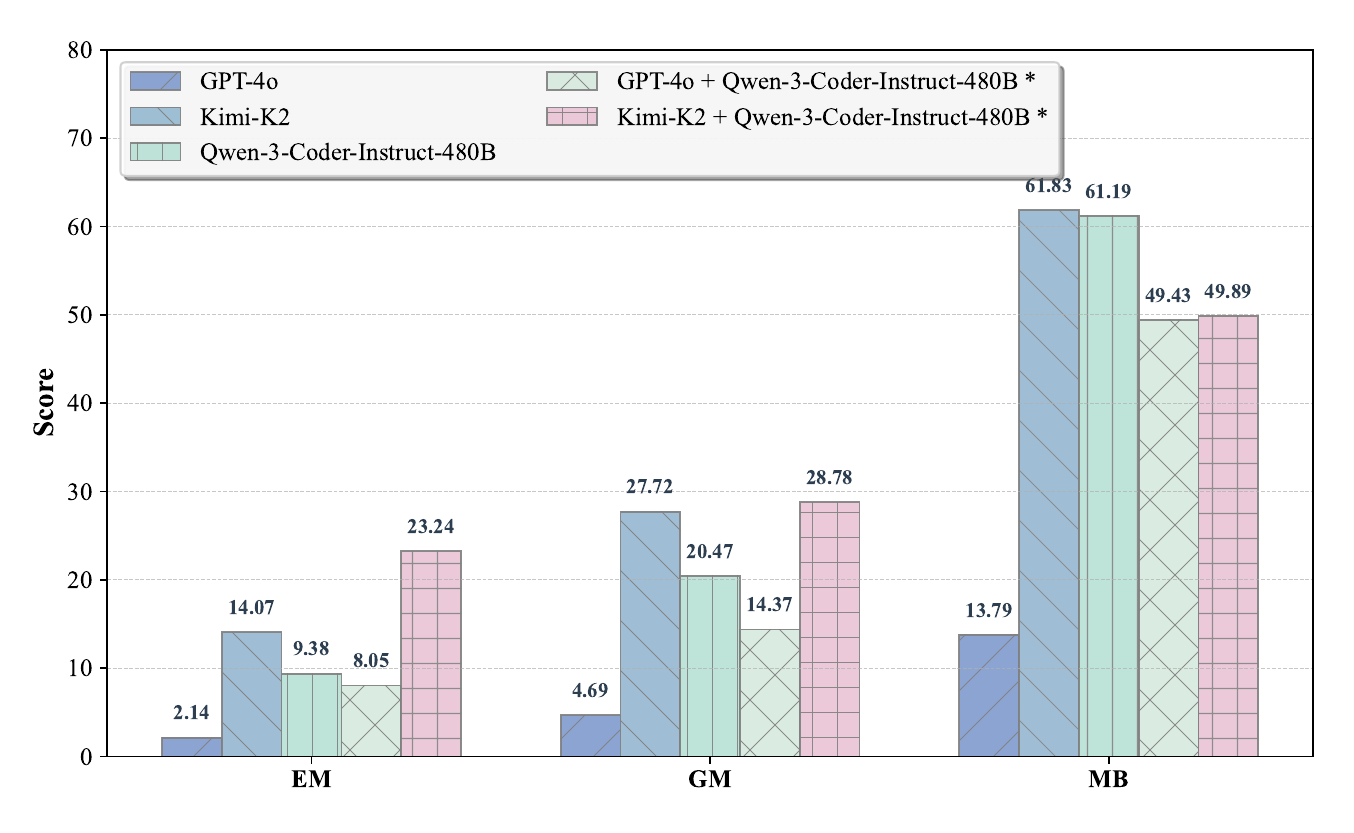}
    \caption{Agent performance on \ourbenchsyn. `$*$' denotes agent-based methods, while others are single-model baselines.}
    \label{fig:agent perfomance comparsion}
\end{figure}

As shown in Appendix~\ref{app:agent baseline} and Figure~\ref{fig:agent}, we introduce an agentic baseline where a main agent plans SQL fixes and a sub-agent implements them, iterating with TQS\footnote{The TQS tool is introduced in Appendix~\ref {app: tqs}.} feedback until completion. Figure~\ref{fig:agent perfomance comparsion} shows that \textbf{agent-based systems can significantly boost performance, but results heavily depend on the main agent’s capabilities.} For example, using Kimi-K2 as the main agent and Qwen3-Coder as the sub-agent increases EM accuracy by 65\% compared to the Kimi-K2 single-model baseline. In contrast, when GPT-4o serves as the main agent—despite a 300\%+ gain over its single-model performance—the combined system still underperforms the single Qwen3-Coder model. We also observe a decline in the MB score of agent-based systems, as multiple rounds of modification can gradually cause the model to deviate from the original SQL. These observations provide initial insights for future exploration of agentic methods in SQL debugging.

\section{Related Work}
The landscape of modern software engineering is increasingly shaped by the integration of LLMs to automate and augment developer tasks.

\paragraph{Code Generation and Text-to-SQL Benchmarks.}
Early text-to-code benchmarks, including HumanEval~\citep{chen2021humaneval}, SQL-Spider~\citep{yu2018spider}, and BIRD~\citep{li2024bird}, focus on simple and short code snippets~\citep{zhuo2025bigcodebench,jain2025livecodebench,fullstackbench}. To address the gap with real-world applications, SWE-Bench~\citep{swebench} evaluates models on complete software issues, which require a comprehensive understanding of codebases. Similarly, Spider2.0~\citep{lei2025spider} extends Text-to-SQL evaluation to enterprise contexts. BIRD-Critic~\citep{li2024bird} introduces SQL debugging, but it only handles short, simplified StackOverflow queries that lack enterprise-level complexity. Most of these benchmarks rely on manually curated datasets, which are costly and prone to data leakage~\citep{chou2025autocodebench}. In this work, we introduce the first enterprise-level \task benchmark, which is automatically constructed via reverse engineering.

\paragraph{LLMs for Automated Software Engineering.}
Recent work applies LLMs to automated software engineering through three primary paradigms: 
(1) \textbf{Single-model} approaches, which attempt to produce patches directly from a description and buggy code, often using few-shot prompting or SFT~\citep{10.1109/ASE56229.2023.00181,yasunaga2021break,allamanis2021self}. These single-model methods are bottlenecked by the need to build large-scale SFT datasets~\citep{pan2024swegym,bird-critic,ma2024lingmaswegptopendevelopmentprocesscentric,yang2025swesmith,pham2025swesyn}.
(2) \textbf{Multi-stage Workflows}, which guide models through defect localization, patch generation, and validation~\citep{xia2024agentless,zhang2024autocoderover,gong2025sqlens}. 
(3) \textbf{Agent-based Methods}, which leverage analysis, execution traces, or test feedback for iterative refinement~\citep{yang2024sweagent,wang2025openhands,bouzenia2024repairagent,chen2023teaching}.
In this work, we provide both SFT-based Single-model solutions and Agent-based methods, offering the community a comprehensive understanding of SQL debugging tasks.

\section{Conclusion}
We introduce \ourbench, the first benchmark for enterprise-level SQL debugging. With its automated construction workflow and execution-free evaluation, \ourbench enables scalable and reliable assessment of LLMs. Despite recent advances in LLM reasoning, our evaluation of nearly 30 models shows that real-world enterprise SQL debugging remains a significant challenge. To encourage further progress, we highlight four promising directions, including three SFT-based strategies and one agent-driven approach. Importantly, \ourbench correlates strongly with practical debugging performance, making it a reliable reference for both academic research and industrial deployment.

\section*{Impact Statement}

This paper presents work whose goal is to advance the field
of Machine Learning. There are many potential societal
consequences of our work, none of which we feel must be
specifically highlighted here.

\bibliography{iclr2026_conference}

@inproceedings{Begoli_2018, series={SIGMOD/PODS ’18},
   title={Apache Calcite: A Foundational Framework for Optimized Query Processing Over Heterogeneous Data Sources},
   url={http://dx.doi.org/10.1145/3183713.3190662},
   DOI={10.1145/3183713.3190662},
   booktitle={Proceedings of the 2018 International Conference on Management of Data},
   publisher={ACM},
   author={Begoli, Edmon and Camacho-Rodríguez, Jesús and Hyde, Julian and Mior, Michael J. and Lemire, Daniel},
   year={2018},
   month=may, pages={221–230},
   collection={SIGMOD/PODS ’18} }

@inproceedings{
gong2025sqlens,
title={{SQL}ens: An End-to-End Framework for Error Detection and Correction in Text-to-{SQL}},
author={Yue Gong and Chuan Lei and Xiao Qin and Kapil Vaidya and Balakrishnan Murali Narayanaswamy and Tim Kraska},
booktitle={The Thirty-ninth Annual Conference on Neural Information Processing Systems},
year={2026},
url={https://openreview.net/forum?id=on6Hf0KP20}
}

@inproceedings{
chou2025autocodebench,
title={AutoCodeBench: Large Language Models are Automatic Code Benchmark Generators},
author={Changzhi Zhou and Ao Liu and Yuchi Deng and Zhiying Zeng and Tao Zhang and Haotian Zhu and Jianwei Cai and Yue Mao and Chenchen Zhang and Lingyun Tan and ZiyanXU and Bohui Zhai and HengyiLIu and Speed Zhu and Wiggin Zhou and Fengzong Lian},
booktitle={The Fourteenth International Conference on Learning Representations},
year={2026},
url={https://openreview.net/forum?id=fN0MED2Idq}
}

@inproceedings{zheng2024llamafactory,
    title = "{L}lama{F}actory: Unified Efficient Fine-Tuning of 100+ Language Models",
    author = "Zheng, Yaowei  and
      Zhang, Richong  and
      Zhang, Junhao  and
      Ye, Yanhan  and
      Luo, Zheyan",
    editor = "Cao, Yixin  and
      Feng, Yang  and
      Xiong, Deyi",
    booktitle = "Proceedings of the 62nd Annual Meeting of the Association for Computational Linguistics (Volume 3: System Demonstrations)",
    month = aug,
    year = "2024",
    address = "Bangkok, Thailand",
    publisher = "Association for Computational Linguistics",
    url = "https://aclanthology.org/2024.acl-demos.38/",
    doi = "10.18653/v1/2024.acl-demos.38",
    pages = "400--410"
}

@inproceedings{kwon2023efficient,
  title={Efficient Memory Management for Large Language Model Serving with PagedAttention},
  author={Woosuk Kwon and Zhuohan Li and Siyuan Zhuang and Ying Sheng and Lianmin Zheng and Cody Hao Yu and Joseph E. Gonzalez and Hao Zhang and Ion Stoica},
  booktitle={Proceedings of the ACM SIGOPS 29th Symposium on Operating Systems Principles},
  year={2023}
}

@inproceedings{zhan-etal-2025-towards,
    title = "Towards Database-Free Text-to-{SQL} Evaluation: A Graph-Based Metric for Functional Correctness",
    author = "Zhan, Yi  and
      Cui, Longjie  and
      Weng, Han  and
      Wang, Guifeng  and
      Tian, Yu  and
      Liu, Boyi  and
      Yang, Yingxiang  and
      Yin, Xiaoming  and
      Xie, Jiajun  and
      Sun, Yang",
    editor = "Rambow, Owen  and
      Wanner, Leo  and
      Apidianaki, Marianna  and
      Al-Khalifa, Hend  and
      Eugenio, Barbara Di  and
      Schockaert, Steven",
    booktitle = "Proceedings of the 31st International Conference on Computational Linguistics",
    month = jan,
    year = "2025",
    address = "Abu Dhabi, UAE",
    publisher = "Association for Computational Linguistics",
    url = "https://aclanthology.org/2025.coling-main.308/",
    pages = "4586--4610"
}

@misc{gpt-5,
      title={Introducing GPT-5}, 
      author={Openai},
      year={2025},
      url={https://openai.com/index/introducing-gpt-5/}, 
}

@misc{deepseek-v3.1,
      title={DeepSeek-V3.1}, 
      author={DeepSeek},
      year={2025},
      url={https://api-docs.deepseek.com/news/news250821}, 
}

@misc{claude,
      title={Claude}, 
      author={anthropic},
      year={2025},
      url={https://www.anthropic.com/claude/sonnet}, 
}

@misc{yang2025qwen3,
      title={Qwen3 Technical Report}, 
      author={An Yang and Anfeng Li and Baosong Yang and Beichen Zhang and Binyuan Hui and Bo Zheng and Bowen Yu and Chang Gao and Chengen Huang and Chenxu Lv and Chujie Zheng and Dayiheng Liu and Fan Zhou and Fei Huang and Feng Hu and Hao Ge and Haoran Wei and Huan Lin and Jialong Tang and Jian Yang and Jianhong Tu and Jianwei Zhang and Jianxin Yang and Jiaxi Yang and Jing Zhou and Jingren Zhou and Junyang Lin and Kai Dang and Keqin Bao and Kexin Yang and Le Yu and Lianghao Deng and Mei Li and Mingfeng Xue and Mingze Li and Pei Zhang and Peng Wang and Qin Zhu and Rui Men and Ruize Gao and Shixuan Liu and Shuang Luo and Tianhao Li and Tianyi Tang and Wenbiao Yin and Xingzhang Ren and Xinyu Wang and Xinyu Zhang and Xuancheng Ren and Yang Fan and Yang Su and Yichang Zhang and Yinger Zhang and Yu Wan and Yuqiong Liu and Zekun Wang and Zeyu Cui and Zhenru Zhang and Zhipeng Zhou and Zihan Qiu},
      year={2025},
      eprint={2505.09388},
      archivePrefix={arXiv},
      primaryClass={cs.CL},
      url={https://arxiv.org/abs/2505.09388}, 
}

@misc{seed1.6,
  title = {Introduction to Techniques Used in Seed1.6},
  author = {{Seed}},
  url = {https://seed.bytedance.com/en/seed1_6},
  year = {2025}
}

@misc{gpt4o,
  title = {Hello {GPT-4o}},
  author = {{OpenAI}},
  url = {https://openai.com/index/hello-gpt-4o/},
  year = {2024}
}

@misc{GPT-4.1,
  title = {Introducing GPT-4.1 in the API},
  author = {{OpenAI}},
  url = {https://openai.com/index/gpt-4-1/},
  year = {2025}
}

@misc{qwen3coder,
  title = {Qwen3-Coder: Agentic Coding in the World},
  author = {{Qwen}},
  url = {https://qwenlm.github.io/blog/qwen3-coder/},
  year = {2025}
}

@misc{gemini2.5,
      title={Gemini 2.5: Pushing the Frontier with Advanced Reasoning, Multimodality, Long Context, and Next Generation Agentic Capabilities}, 
      author={Gemini},
      year={2025},
      eprint={2507.06261},
      archivePrefix={arXiv},
      primaryClass={cs.CL},
      url={https://arxiv.org/abs/2507.06261}, 
}

@misc{deepseekai2025deepseekr1,
      title={DeepSeek-R1: Incentivizing Reasoning Capability in LLMs via Reinforcement Learning}, 
      author={DeepSeek-AI},
      year={2025},
      eprint={2501.12948},
      archivePrefix={arXiv},
      primaryClass={cs.CL},
      url={https://arxiv.org/abs/2501.12948}, 
}

@misc{o3,
  title = {Introducing OpenAI o3 and o4-mini},
  author = {OpenAI},
  url = {https://openai.com/index/introducing-o3-and-o4-mini/},
  year={2025}
}

@misc{claude4,
  title = {Introducing Claude 4},
  author = {Anthropic},
  url = {https://www.anthropic.com/news/claude-4},
  year={2025}
}

@misc{hui2024qwen25codertechnicalreport,
      title={Qwen2.5-Coder Technical Report}, 
      author={Binyuan Hui and Jian Yang and Zeyu Cui and Jiaxi Yang and Dayiheng Liu and Lei Zhang and Tianyu Liu and Jiajun Zhang and Bowen Yu and Keming Lu and Kai Dang and Yang Fan and Yichang Zhang and An Yang and Rui Men and Fei Huang and Bo Zheng and Yibo Miao and Shanghaoran Quan and Yunlong Feng and Xingzhang Ren and Xuancheng Ren and Jingren Zhou and Junyang Lin},
      year={2024},
      eprint={2409.12186},
      archivePrefix={arXiv},
      primaryClass={cs.CL},
      url={https://arxiv.org/abs/2409.12186}, 
}

@misc{deepseek_v3,
      title={DeepSeek-V3 Technical Report}, 
      author={DeepSeek-AI},
      year={2025},
      eprint={2412.19437},
      archivePrefix={arXiv},
      primaryClass={cs.CL},
      url={https://arxiv.org/abs/2412.19437}, 
}

@misc{kimiteam2025kimik2openagentic,
      title={Kimi K2: Open Agentic Intelligence}, 
      author={Kimi-Team},
      year={2025},
      eprint={2507.20534},
      archivePrefix={arXiv},
      primaryClass={cs.LG},
      url={https://arxiv.org/abs/2507.20534}, 
}

@misc{fullstackbench,
      title={FullStack Bench: Evaluating LLMs as Full Stack Coders}, 
      author={Siyao Liu and He Zhu and Jerry Liu and Shulin Xin and Aoyan Li and Rui Long and Li Chen and Jack Yang and Jinxiang Xia and Z. Y. Peng and Shukai Liu and Zhaoxiang Zhang and Ge Zhang and Wenhao Huang and Kai Shen and Liang Xiang},
      year={2024},
      eprint={2412.00535},
      archivePrefix={arXiv},
      primaryClass={cs.AI},
      url={https://arxiv.org/abs/2412.00535}, 
}

@inproceedings{
jain2025livecodebench,
title={LiveCodeBench: Holistic and Contamination Free Evaluation of Large Language Models for Code},
author={Naman Jain and King Han and Alex Gu and Wen-Ding Li and Fanjia Yan and Tianjun Zhang and Sida Wang and Armando Solar-Lezama and Koushik Sen and Ion Stoica},
booktitle={The Thirteenth International Conference on Learning Representations},
year={2025},
url={https://openreview.net/forum?id=chfJJYC3iL}
}

@inproceedings{
zhuo2025bigcodebench,
title={BigCodeBench: Benchmarking Code Generation with Diverse Function Calls and Complex Instructions},
author={Terry Yue Zhuo and Vu Minh Chien and Jenny Chim and Han Hu and Wenhao Yu and Ratnadira Widyasari and Imam Nur Bani Yusuf and Haolan Zhan and Junda He and Indraneil Paul and Simon Brunner and Chen GONG and James Hoang and Armel Randy Zebaze and Xiaoheng Hong and Wen-Ding Li and Jean Kaddour and Ming Xu and Zhihan Zhang and Prateek Yadav and Naman Jain and Alex Gu and Zhoujun Cheng and Jiawei Liu and Qian Liu and Zijian Wang and David Lo and Binyuan Hui and Niklas Muennighoff and Daniel Fried and Xiaoning Du and Harm de Vries and Leandro Von Werra},
booktitle={The Thirteenth International Conference on Learning Representations},
year={2025},
url={https://openreview.net/forum?id=YrycTjllL0}
}

@inproceedings{
laban2025llmslostmultiturnconversation,
title={{LLM}s Get Lost In Multi-Turn Conversation},
author={Philippe Laban and Hiroaki Hayashi and Yingbo Zhou and Jennifer Neville},
booktitle={The Fourteenth International Conference on Learning Representations},
year={2026},
url={https://openreview.net/forum?id=VKGTGGcwl6}
}

@inproceedings{duan2024pdcdmsftroad,
    title = "{PDC} {\&} {DM}-{SFT}: A Road for {LLM} {SQL} Bug-Fix Enhancing",
    author = "Duan, Yiwen  and
      Yu, Yonghong  and
      Zhao, Xiaoming  and
      Wu, Yichang  and
      Liu, Wenbo",
    editor = "Rambow, Owen  and
      Wanner, Leo  and
      Apidianaki, Marianna  and
      Al-Khalifa, Hend  and
      Eugenio, Barbara Di  and
      Schockaert, Steven  and
      Darwish, Kareem  and
      Agarwal, Apoorv",
    booktitle = "Proceedings of the 31st International Conference on Computational Linguistics: Industry Track",
    month = jan,
    year = "2025",
    address = "Abu Dhabi, UAE",
    publisher = "Association for Computational Linguistics",
    url = "https://aclanthology.org/2025.coling-industry.7/",
    pages = "76--90"
}

@inproceedings{bouzenia2025understandingsoftwareengineeringagents,
  title={Understanding Software Engineering Agents: A Study of Thought-Action-Result Trajectories},
  author={Bouzenia, Islem and Pradel, Michael},
  booktitle={2025 40th IEEE/ACM International Conference on Automated Software Engineering (ASE)},
  pages={2846--2857},
  year={2025},
  organization={IEEE}
}

@misc{seed2025seedcoderletcodemodel,
      title={Seed-Coder: Let the Code Model Curate Data for Itself}, 
      author={ByteDance Seed and Yuyu Zhang and Jing Su and Yifan Sun and Chenguang Xi and Xia Xiao and Shen Zheng and Anxiang Zhang and Kaibo Liu and Daoguang Zan and Tao Sun and Jinhua Zhu and Shulin Xin and Dong Huang and Yetao Bai and Lixin Dong and Chao Li and Jianchong Chen and Hanzhi Zhou and Yifan Huang and Guanghan Ning and Xierui Song and Jiaze Chen and Siyao Liu and Kai Shen and Liang Xiang and Yonghui Wu},
      year={2025},
      eprint={2506.03524},
      archivePrefix={arXiv},
      primaryClass={cs.CL},
      url={https://arxiv.org/abs/2506.03524}, 
}

@article{li2024bird,
  title={Can llm already serve as a database interface? a big bench for large-scale database grounded text-to-sqls},
  author={Li, Jinyang and Hui, Binyuan and Qu, Ge and Yang, Jiaxi and Li, Binhua and Li, Bowen and Wang, Bailin and Qin, Bowen and Geng, Ruiying and Huo, Nan and others},
  journal={Advances in Neural Information Processing Systems},
  volume={36},
  year={2024}
}

@inproceedings{chamberlin1974,
  author    = {Donald D. Chamberlin and Raymond F. Boyce},
  title     = {{SEQUEL}: A Structured English Query Language},
  booktitle = {Proceedings of the 1974 ACM SIGMOD Workshop on Data Description, Access and Control},
  year      = {1974},
  pages     = {249--264},
  publisher = {ACM}
}

@inproceedings{armbrust2015,
  author       = {Michael Armbrust and
                  Reynold S. Xin and
                  Cheng Lian and
                  Yin Huai and
                  Davies Liu and
                  Joseph K. Bradley and
                  Xiangrui Meng and
                  Tomer Kaftan and
                  Michael J. Franklin and
                  Ali Ghodsi and
                  Matei Zaharia},
  editor       = {Timos K. Sellis and
                  Susan B. Davidson and
                  Zachary G. Ives},
  title        = {Spark {SQL:} Relational Data Processing in Spark},
  booktitle    = {Proceedings of the 2015 {ACM} {SIGMOD} International Conference on
                  Management of Data, Melbourne, Victoria, Australia, May 31 - June
                  4, 2015},
  pages        = {1383--1394},
  publisher    = {{ACM}},
  year         = {2015},
}

@misc{pan2024swegym,
      title={Training Software Engineering Agents and Verifiers with SWE-Gym}, 
      author={Jiayi Pan and Xingyao Wang and Graham Neubig and Navdeep Jaitly and Heng Ji and Alane Suhr and Yizhe Zhang},
      year={2024},
      eprint={2412.21139},
      archivePrefix={arXiv},
      primaryClass={cs.SE},
      url={https://arxiv.org/abs/2412.21139}, 
}

@misc{pham2025swesyn,
      title={SWE-Synth: Synthesizing Verifiable Bug-Fix Data to Enable Large Language Models in Resolving Real-World Bugs}, 
      author={Minh V. T. Pham and Huy N. Phan and Hoang N. Phan and Cuong Le Chi and Tien N. Nguyen and Nghi D. Q. Bui},
      year={2025},
      eprint={2504.14757},
      archivePrefix={arXiv},
      primaryClass={cs.SE},
      url={https://arxiv.org/abs/2504.14757}, 
}

@inproceedings{
yang2025swesmith,
title={{SWE}-smith: Scaling Data for Software Engineering Agents},
author={John Yang and Kilian Lieret and Carlos E Jimenez and Alexander Wettig and Kabir Khandpur and Yanzhe Zhang and Binyuan Hui and Ofir Press and Ludwig Schmidt and Diyi Yang},
booktitle={The Thirty-ninth Annual Conference on Neural Information Processing Systems Datasets and Benchmarks Track},
year={2026},
url={https://openreview.net/forum?id=63iVrXc8cC}
}

@misc{ma2024lingmaswegptopendevelopmentprocesscentric,
      title={Lingma SWE-GPT: An Open Development-Process-Centric Language Model for Automated Software Improvement}, 
      author={Yingwei Ma and Rongyu Cao and Yongchang Cao and Yue Zhang and Jue Chen and Yibo Liu and Yuchen Liu and Binhua Li and Fei Huang and Yongbin Li},
      year={2024},
      eprint={2411.00622},
      archivePrefix={arXiv},
      primaryClass={cs.SE},
      url={https://arxiv.org/abs/2411.00622}, 
}

@inproceedings{bouzenia2024repairagent,
author = {Bouzenia, Islem and Devanbu, Premkumar and Pradel, Michael},
title = {RepairAgent: An Autonomous, LLM-Based Agent for Program Repair},
year = {2025},
isbn = {9798331505691},
publisher = {IEEE Press},
url = {https://doi.org/10.1109/ICSE55347.2025.00157},
doi = {10.1109/ICSE55347.2025.00157},
booktitle = {Proceedings of the IEEE/ACM 47th International Conference on Software Engineering},
pages = {2188–2200},
numpages = {13},
location = {Ottawa, Ontario, Canada},
series = {ICSE '25}
}

@article{allamanis2021self,
  title={Self-supervised bug detection and repair},
  author={Allamanis, Miltiadis and Jackson-Flux, Henry and Brockschmidt, Marc},
  journal={Advances in Neural Information Processing Systems},
  volume={34},
  pages={27865--27876},
  year={2021}
}

@inproceedings{yasunaga2021break,
  title={Break-it-fix-it: Unsupervised learning for program repair},
  author={Yasunaga, Michihiro and Liang, Percy},
  booktitle={International conference on machine learning},
  pages={11941--11952},
  year={2021},
  organization={PMLR}
}

@inproceedings{10.1109/ASE56229.2023.00181,
author = {Huang, Kai and Meng, Xiangxin and Zhang, Jian and Liu, Yang and Wang, Wenjie and Li, Shuhao and Zhang, Yuqing},
title = {An Empirical Study on Fine-Tuning Large Language Models of Code for Automated Program Repair},
year = {2024},
isbn = {9798350329964},
publisher = {IEEE Press},
url = {https://doi.org/10.1109/ASE56229.2023.00181},
doi = {10.1109/ASE56229.2023.00181},
booktitle = {Proceedings of the 38th IEEE/ACM International Conference on Automated Software Engineering},
pages = {1162–1174},
numpages = {13},
location = {Echternach, Luxembourg},
series = {ASE '23}
}

@inproceedings{yu2018spider,
  title={Spider: A Large-Scale Human-Labeled Dataset for Complex and Cross-Domain Semantic Parsing and Text-to-SQL Task},
  author={Yu, Tao and Zhang, Rui and Yang, Kai and Yasunaga, Michihiro and Wang, Dongxu and Li, Zifan and Ma, James and Li, Irene and Yao, Qingning and Roman, Shanelle and others},
  booktitle={Proceedings of the 2018 Conference on Empirical Methods in Natural Language Processing},
  pages={3911--3921},
  year={2018}
}

@misc{
zhong2017wikisql,
title={Seq2{SQL}: Generating Structured Queries From Natural Language Using Reinforcement Learning },
author={Victor Zhong and Caiming Xiong and Richard Socher},
year={2018},
url={https://openreview.net/forum?id=Syx6bz-Ab},
}

@inproceedings{yang2024sweagent,
author = {Yang, John and Jimenez, Carlos E. and Wettig, Alexander and Lieret, Kilian and Yao, Shunyu and Narasimhan, Karthik and Press, Ofir},
title = {SWE-agent: agent-computer interfaces enable automated software engineering},
year = {2024},
isbn = {9798331314385},
publisher = {Curran Associates Inc.},
address = {Red Hook, NY, USA},
booktitle = {Proceedings of the 38th International Conference on Neural Information Processing Systems},
articleno = {1601},
numpages = {125},
location = {Vancouver, BC, Canada},
series = {NIPS '24}
}

@misc{CoderR,
      title={CodeR: Issue Resolving with Multi-Agent and Task Graphs}, 
      author={Dong Chen and Shaoxin Lin and Muhan Zeng and Daoguang Zan and Jian-Gang Wang and Anton Cheshkov and Jun Sun and Hao Yu and Guoliang Dong and Artem Aliev and Jie Wang and Xiao Cheng and Guangtai Liang and Yuchi Ma and Pan Bian and Tao Xie and Qianxiang Wang},
      year={2024},
      eprint={2406.01304},
      archivePrefix={arXiv},
      primaryClass={cs.CL},
      url={https://arxiv.org/abs/2406.01304}, 
}

@article{xia2024agentless,
author = {Xia, Chunqiu Steven and Deng, Yinlin and Dunn, Soren and Zhang, Lingming},
title = {Demystifying LLM-Based Software Engineering Agents},
year = {2025},
issue_date = {July 2025},
publisher = {Association for Computing Machinery},
address = {New York, NY, USA},
volume = {2},
number = {FSE},
url = {https://doi.org/10.1145/3715754},
doi = {10.1145/3715754},
journal = {Proc. ACM Softw. Eng.},
month = jun,
articleno = {FSE037},
numpages = {24}
}

@misc{chen2021humaneval,
      title={Evaluating Large Language Models Trained on Code}, 
      author={Mark Chen and Jerry Tworek and Heewoo Jun and Qiming Yuan and Henrique Ponde de Oliveira Pinto and Jared Kaplan and Harri Edwards and Yuri Burda and Nicholas Joseph and Greg Brockman and Alex Ray and Raul Puri and Gretchen Krueger and Michael Petrov and Heidy Khlaaf and Girish Sastry and Pamela Mishkin and Brooke Chan and Scott Gray and Nick Ryder and Mikhail Pavlov and Alethea Power and Lukasz Kaiser and Mohammad Bavarian and Clemens Winter and Philippe Tillet and Felipe Petroski Such and Dave Cummings and Matthias Plappert and Fotios Chantzis and Elizabeth Barnes and Ariel Herbert-Voss and William Hebgen Guss and Alex Nichol and Alex Paino and Nikolas Tezak and Jie Tang and Igor Babuschkin and Suchir Balaji and Shantanu Jain and William Saunders and Christopher Hesse and Andrew N. Carr and Jan Leike and Josh Achiam and Vedant Misra and Evan Morikawa and Alec Radford and Matthew Knight and Miles Brundage and Mira Murati and Katie Mayer and Peter Welinder and Bob McGrew and Dario Amodei and Sam McCandlish and Ilya Sutskever and Wojciech Zaremba},
      year={2021},
      eprint={2107.03374},
      archivePrefix={arXiv},
      primaryClass={cs.LG},
      url={https://arxiv.org/abs/2107.03374}, 
}

@inproceedings{zhang2024autocoderover,
  title={Autocoderover: Autonomous program improvement},
  author={Zhang, Yuntong and Ruan, Haifeng and Fan, Zhiyu and Roychoudhury, Abhik},
  booktitle={Proceedings of the 33rd ACM SIGSOFT International Symposium on Software Testing and Analysis},
  pages={1592--1604},
  year={2024}
}

@inproceedings{chen2023teaching,
title={Teaching Large Language Models to Self-Debug},
author={Xinyun Chen and Maxwell Lin and Nathanael Sch{\"a}rli and Denny Zhou},
booktitle={The Twelfth International Conference on Learning Representations},
year={2024},
}

@inproceedings{wang2025openhands,
title={OpenHands: An Open Platform for {AI} Software Developers as Generalist Agents},
author={Xingyao Wang and Boxuan Li and Yufan Song and Frank F. Xu and Xiangru Tang and Mingchen Zhuge and Jiayi Pan and Yueqi Song and Bowen Li and Jaskirat Singh and Hoang H. Tran and Fuqiang Li and Ren Ma and Mingzhang Zheng and Bill Qian and Yanjun Shao and Niklas Muennighoff and Yizhe Zhang and Binyuan Hui and Junyang Lin and Robert Brennan and Hao Peng and Heng Ji and Graham Neubig},
booktitle={The Thirteenth International Conference on Learning Representations},
year={2025},
}

@inproceedings{lei2025spider,
title={Spider 2.0: Evaluating Language Models on Real-World Enterprise Text-to-{SQL} Workflows},
author={Fangyu Lei and Jixuan Chen and Yuxiao Ye and Ruisheng Cao and Dongchan Shin and Hongjin SU and ZHAOQING SUO and Hongcheng Gao and Wenjing Hu and Pengcheng Yin and Victor Zhong and Caiming Xiong and Ruoxi Sun and Qian Liu and Sida Wang and Tao Yu},
booktitle={The Thirteenth International Conference on Learning Representations},
year={2025},
}

@inproceedings{
bird-critic,
title={{SWE}-{SQL}: Illuminating {LLM} Pathways to Solve User {SQL} Issues in Real-World Applications},
author={Jinyang Li and Xiaolong Li and Ge Qu and Per Jacobsson and Bowen Qin and Binyuan Hui and Shuzheng Si and Nan Huo and Xiaohan Xu and Yue Zhang and Ziwei Tang and Yuanshuai Li and Florensia Widjaja and Xintong Zhu and Feige Zhou and Yongfeng Huang and Yannis Papakonstantinou and Fatma Ozcan and Chenhao Ma and Reynold Cheng},
booktitle={The Thirty-ninth Annual Conference on Neural Information Processing Systems},
year={2026},
url={https://openreview.net/forum?id=yRxXTdElLv}
}

@article{li2025omnisql,
author = {Li, Haoyang and Wu, Shang and Zhang, Xiaokang and Huang, Xinmei and Zhang, Jing and Jiang, Fuxin and Wang, Shuai and Zhang, Tieying and Chen, Jianjun and Shi, Rui and Chen, Hong and Li, Cuiping},
title = {OmniSQL: Synthesizing High-Quality Text-to-SQL Data at Scale},
year = {2025},
issue_date = {July 2025},
publisher = {VLDB Endowment},
volume = {18},
number = {11},
issn = {2150-8097},
url = {https://doi.org/10.14778/3749646.3749723},
doi = {10.14778/3749646.3749723},
journal = {Proc. VLDB Endow.},
month = jul,
pages = {4695–4709},
numpages = {15}
}

@inproceedings{
swebench,
title={{SWE}-bench: Can Language Models Resolve Real-world Github Issues?},
author={Carlos E Jimenez and John Yang and Alexander Wettig and Shunyu Yao and Kexin Pei and Ofir Press and Karthik R Narasimhan},
booktitle={The Twelfth International Conference on Learning Representations},
year={2024},
}
\bibliographystyle{icml2026}

\newpage
\appendix
\onecolumn
\clearpage
\section*{Appendix}

\section{Use of LLMs}
\label{app:use of LLMs}
We disclose the following uses of LLMs in this work:
\begin{enumerate}
\item \textbf{Data Construction.} LLMs were used during benchmark and dataset construction. Details of the data generation process are provided in Section~\ref{sec:benchmark}, and the corresponding prompts are included in Appendix~\ref{sec:prompts}.
\item \textbf{Benchmark Evaluation.} LLMs were evaluated on the benchmark introduced in this paper. The evaluated models are listed in Appendix~\ref{app:llms}.
\item \textbf{Manuscript Preparation.} LLMs were used exclusively for language editing, proofreading, and stylistic refinement of the manuscript.
\item \textbf{Programming Assistance.} LLMs were used to assist with software development. All generated code was manually reviewed and validated by the authors.
\end{enumerate}

All research ideas, methodological designs, experimental decisions, analyses, and scientific conclusions presented in this work were conceived and developed solely by the authors. No LLM contributed to the intellectual content or scientific claims of this research.

\section{Background of ETL SQL debugging.}
\label{app:background}
Our benchmark targets industrial Extract–Transform–Load (ETL) workloads, which differ substantially from traditional Text-to-SQL or analytics-oriented SQL generation. We summarize the key distinctions below.

\paragraph{Task objectives and nature.}
ETL is primarily a data engineering task focused on preparing, transforming, and integrating raw data into a clean and consistent warehouse for downstream consumption. The goal is reliable data production rather than interactive analysis.
In contrast, Data Analysis / Text-to-SQL tasks aim to explore existing datasets and answer analytical questions. These tasks resemble the work of data analysts: flexible, insight-driven, and focused on extracting knowledge from already-curated data rather than producing new datasets.

\paragraph{Data scale and operations.}
ETL pipelines operate on full-scale production data. For example, computing daily active users may require scanning and joining hundreds of millions of raw event records. The dominant SQL operations involve \texttt{INSERT}, \texttt{UPDATE}, \texttt{DELETE}, and \texttt{MERGE}, reflecting an emphasis on data movement, reshaping, and materialization.
By contrast, Data Analysis / Text-to-SQL workloads typically query curated warehouse tables using complex \texttt{SELECT} statements that return relatively small result sets—reports, leaderboards, or summary statistics. These tasks focus on the correctness of the query output rather than on large-scale data transformation.




\section{Source of SQL Bug Taxonomy.}
To construct the SQL bug taxonomy, we analyze a corpus of production logs and select 268 representative samples for detailed manual inspection. Three domain experts—each with more than three years of professional experience in SQL analysis and data quality verification—annotate these logs over a two-week period. The error categories identified in these annotations form the basis of the final SQL Bug Taxonomy.

\section{Experimental Settings}
\label{app:experimental settings}

\subsection{Evaluation}
\label{app: evaluation}

\subsubsection{Challenges in Execution-Based Evaluation}
Evaluating enterprise SQL generation and debugging systems presents several unique challenges.
\textit{First}, conventional execution-based accuracy—where correctness is determined by comparing a program’s output to a reference—is often impractical in production environments. This is due to two primary constraints: 
(1) \textbf{Security and Privacy}: Production databases typically contain proprietary or sensitive data, making arbitrary code execution infeasible; 
(2) \textbf{Efficiency}: Executing complex SQL scripts on large-scale production datasets is computationally expensive and time-consuming. 
\textit{Second}, correctness is not binary. Unlike in standardized benchmarks, real-world SQL debugging admits multiple valid solutions. A repair can be correct through various syntactic paths or logical approaches.
\textit{Third}, string-based metrics are a poor proxy for quality. Comparing predicted SQL to a reference string ignores functional equivalence.
Therefore, a robust evaluation framework must balance efficiency and accuracy while reliably reflecting SQL quality in real-world problem-solving.

\subsubsection{Evaluation Metrics}
\label{app: evaluation metrics}

To address challenges in SQL debugging evaluation, we introduce an execution-free evaluation methodology based on three complementary metrics.

\paragraph{Exact Match Score (EM).} 
This metric provides a strict, reproducible measure of syntactic correctness by comparing the predicted SQL string directly against the reference:
\begin{equation}
\text{EM} = \frac{1}{N} \sum_{i=1}^{N} \mathbf{1}\!\left[\hat{q}_i = q_i\right]
\end{equation}
where $\hat{q}_i$ is the predicted SQL, $q_i$ is the reference SQL, and $\mathbf{1}[\cdot]$ is the indicator function. While stringent, EM serves as a clear lower bound on model performance.

\begin{figure}[t]
    \centering
    \includegraphics[width=1\linewidth]{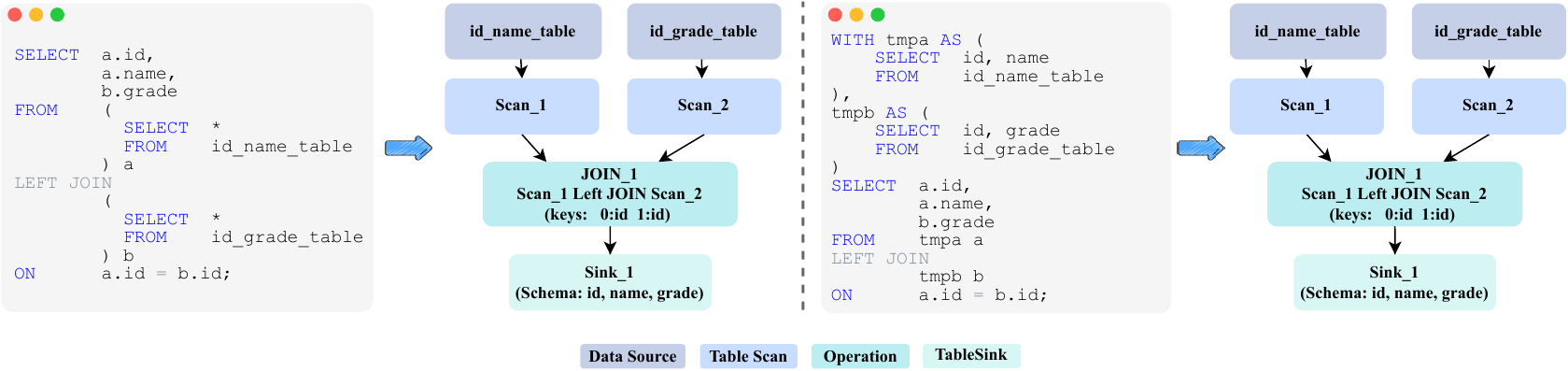}
    \caption{Illustration of Graph Match Score. Although the left and right SQL snippets differ syntactically, their optimized abstract syntax trees are structurally identical. Graph matching evaluates semantic equivalence through tree isomorphism.}
    \label{fig:gm}
\end{figure}

\paragraph{Graph Match Score (GM).}
To assess the semantic equivalence of SQL queries, we represent each query as an optimized abstract syntax tree, illustrated in Figure \ref{fig:gm}. Each node corresponds to a logical relational operator (e.g., \textit{Join, Project, Filter}), and the hierarchical structure encodes operator dependencies and execution order.
We use Apache Calcite \citep{Begoli_2018} to compile SQL queries into its canonical intermediate representation. Calcite applies a suite of logical rewrites—such as operator reordering and clause simplification—to produce normalized logical plans that are robust to superficial syntactic differences.
This intermediate form also encodes both control and data dependencies as edges, yielding graph structures that capture deeper aspects of query semantics. These enriched graphs enable more faithful comparison and interpretation of SQL behavior.
The GM score is computed by performing graph isomorphism over the normalized representations. This allows our method to detect semantic equivalence even when queries differ substantially in surface form:
\begin{equation}
\text{GM} = \frac{1}{N} \sum_{i=1}^{N} \mathbf{1}\!\left[\text{Graph}(\hat{q}_i) \cong \text{Graph}(q_i)\right]
\end{equation}
where $\cong$ denotes graph isomorphism. This approach recognizes semantically equivalent codes that may differ syntactically. 

\paragraph{Modify Better Score (MB).}
For iterative debugging scenarios, absolute correctness is insufficient; we must measure progressive improvement. The MB metric evaluates whether a prediction moves closer to the correct solution by comparing AST edit distances:
\begin{equation}
\text{MB} = \frac{1}{N} \sum_{i=1}^{N} \mathbf{1}\!\left[d(\hat{q}_i, q_i) < d(b_i, q_i)\right]
\end{equation}
where $d(\cdot,\cdot)$ denotes normalized AST edit distance, $\hat{q}_i$ is the predicted repair, $q_i$ is the reference SQL, and $b_i$ is the original buggy query. This metric specifically assesses a model's capacity for incremental repair in debugging workflows.

Together, these metrics provide a comprehensive evaluation framework that balances efficiency, reproducibility, and semantic understanding while avoiding the practical limitations of execution-based assessment.

\subsubsection{Execution-based Validation}
\label{app: tqs}
The earlier discussion on the impracticality of execution accuracy may appear to conflict with the execution checks referenced in this paper. To clarify, all execution-based validation in our work is strictly non-executive—that is, we do not run SQL scripts against a live engine. Instead, we rely on TQS, an enterprise-grade SQL quality validation tool built on Apache Calcite, which performs comprehensive static analysis, including \textbf{syntax and semantic checks} to ensure scripts are syntactically valid and logically well-formed, as well as \textbf{schema and column validation} to confirm that all referenced tables and fields exist in the physical schema. This static-analysis approach provides rigorous error detection during development while avoiding the practical limitations associated with true execution-based evaluation.

\subsubsection{Reliability of Graph Match Score}
\label{app:reliable of gm}
Graph Match (GM) is a relatively strict metric that can occasionally classify correct outputs as incorrect. To assess its reliability, we conducted a human evaluation on 300 samples, which shows that GM agrees with expert judgments in over 68\% of cases. While such omissions can occur, they tend to affect all models fairly. To further validate GM, we analyzed manual annotations on 100 samples for each of three different models. The results indicate that the model ranking based on GM scores is consistent with the ranking derived from human evaluation. This consistency provides evidence that GM scores are a reasonably reliable metric for comparing model performance.

\subsection{LLMs}
\label{app:llms}

This study ensures a robust evaluation by leveraging a diverse set of LLMs, encompassing both open-source and proprietary architectures to cover a broad range of capabilities.

\textbf{Open-Source Models.} DeepSeek-R1-0528~\citep{deepseekai2025deepseekr1}, DeepSeek-V3-0324~\citep{deepseek_v3}, DeepSeek-V3.1, Qwen-2.5-Coder~\citep{hui2024qwen25codertechnicalreport}, Qwen-3-235B-A22B-Instruct-2507~\citep{yang2025qwen3}, Qwen-3-Coder-480B-A35B-Instruct~\citep{qwen3coder}, QwQ-32B, Seed-Coder-8B~\citep{seed2025seedcoderletcodemodel}, and OmniSQL-32B~\citep{li2025omnisql}.

\textbf{Closed-Source Models.} Claude-Sonnet-4~\citep{claude4}, GPT-4o-mini-2024-07-18, GPT-4o-2024-11-20~\citep{gpt4o}, GPT-4.1~\citep{GPT-4.1}, o3-mini~\citep{o3}, o1-Preview, GPT-5, Gemini 2.5 Pro~\citep{gemini2.5}, Kimi-K2~\citep{kimiteam2025kimik2openagentic}, and Doubao-Seed-1.6, Doubao-Seed-1.6-Flash, and Doubao-Seed-1.6-Thinking~\citep{seed1.6}.

\subsection{Baselines}
\label{app:baseline}

\subsubsection{SFT Baselines}
\label{app:sft baseline}

\begin{figure}[htb]
    \centering
    \includegraphics[width=1\linewidth]{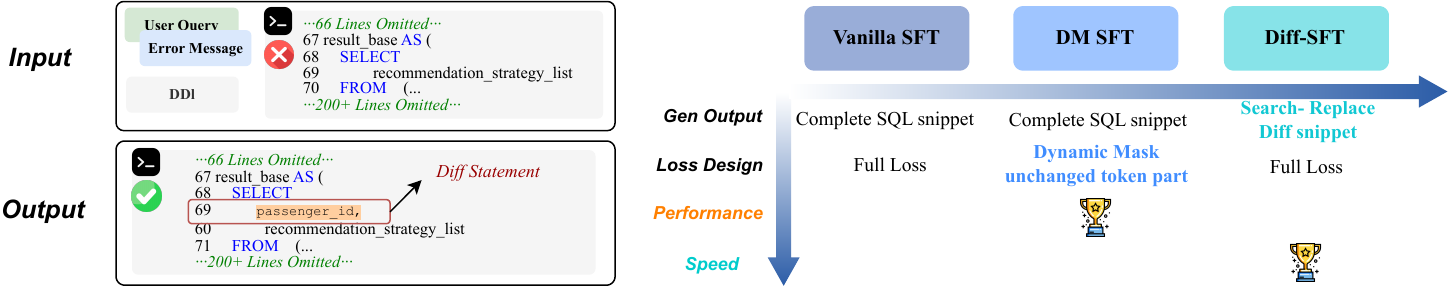}
    \caption{Illustration of three distinct supervised fine-tuning (SFT) methods.}
    \label{fig:sft methods}
\end{figure}

We propose three distinct supervised fine-tuning (SFT) methods as baselines.

\paragraph{Vanilla SFT.} 
This is the standard sequence-to-sequence fine-tuning approach. The model takes as input the error message, the DDL, and the issue SQL, and is trained to generate the complete, corrected reference SQL. While simple, this method establishes a fundamental baseline for performance.

\paragraph{DM-SFT (Dynamic-Masked SFT).} 
In enterprise SQL debugging, the differences between an issue SQL and its reference SQL are often minimal within lengthy code snippets. Consequently, Vanilla SFT models can rapidly reduce loss by learning to copy the large, unchanged portions of the input, potentially failing to focus on the critical, erroneous segments. To mitigate this, we adopt Dynamic-Masked SFT (DM-SFT) \citep{duan2024pdcdmsftroad}, which randomly masks the loss calculation for 50\% of the tokens that are identical between the input and output. By increasing the loss contribution of the changed tokens, this method encourages the model to prioritize learning the necessary edits.

\paragraph{Diff-SFT.} 
Generating the complete SQL code significantly increases inference overhead. We propose an alternative method that outputs only a "diff" snippet, framing the task as a search-and-replace operation. The model's objective is to identify the erroneous code segment in the input and generate the corresponding corrected snippet.

\subsubsection{Agent Baselines}
\label{app:agent baseline}

\begin{figure*}[h]
    \centering
    \includegraphics[width=0.5\linewidth]{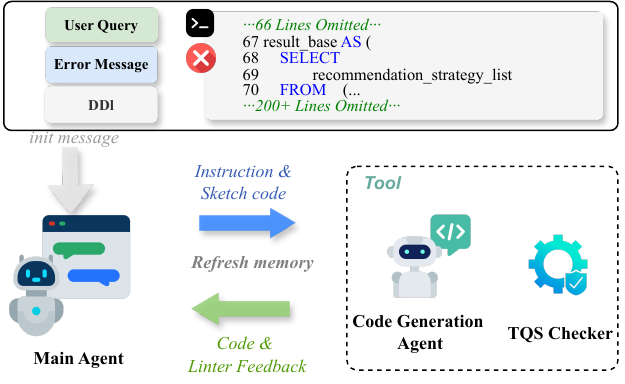}
    \caption{
    Overview of the agentic method, which consists of a main agent, a code-generation sub-agent, and a TQS checker tool. The \textit{main agent} observes the error message and the issued SQL, analyzes the cause of the failure and the required modification, and generates a code-editing instruction for the code generation sub-agent. The \textit{code generation sub-agent} applies the instruction to modify the code; the updated code is automatically passed through a TQS checker to detect errors, and the resulting code and lint feedback are used to update the main agent's memory. This iterative loop continues until the main agent determines that all necessary modifications have been completed.
    }
    \label{fig:agent}
\end{figure*}

As illustrated in Figure~\ref{fig:agent}, we design an \textbf{agentic framework} that coordinates multiple specialized components to iteratively refine generated SQL queries and resolve execution failures. The framework consists of three modules: a main agent, a code-generation sub-agent, and a TQS tool.

The main agent serves as the central controller. It receives the error message and the proposed SQL code for the issue from the previous iteration. Based on these inputs, it analyzes the underlying cause of failure, identifies the modifications required to fix the issue, and produces a structured instruction describing the intended code change. This instruction is sent to the code-generation sub-agent. The code-generation sub-agent performs the actual code editing. It interprets the modification instruction and updates the SQL sketch code accordingly. Once the revision is complete, the generated code is automatically processed by a TQS checker, which detects syntax errors, style violations, and structural inconsistencies. The resulting code and lint feedback are then incorporated into the main agent’s memory. This interaction forms an iterative correction loop. The main agent continuously observes the updated code and diagnostic feedback, issuing refined modification instructions until it concludes that the SQL query is correct and no further edits are required.

\subsection{Dataset}
\label{app:dataset}

Our training dataset consists of three components:
\begin{itemize}[
]
    \item \textbf{Reverse-engineered data}: We manually injected bugs into correct SQL queries collected from production logs, yielding a total of $2,015$ samples.
    \item \textbf{Log-mined data}: We extracted erroneous SQL queries and their associated error messages from online execution logs. For each instance, the reference SQL was manually written and validated by domain experts, resulting in $1,971$ samples.
    \item \textbf{Synthetic data}: We generated additional samples from the BIRD~\citep{li2024bird} and Spider~\citep{yu2018spider} Text-to-SQL datasets to expand the SFT data, producing $1,054$ samples.
\end{itemize}

\subsection{Hyperparameters}
\label{app:hyperparam}

\paragraph{Fine-tuning.} For self-supervised fine-tuning, models are trained for $5$ epoch with a learning rate of $1e-5$ and a per device batch size of $64$. We employed the AdamW optimizer and a cosine learning rate scheduler with a warm-up phase corresponding to $3\%$ of the total training steps. 

\paragraph{Evaluation.} 
Following \citet{yang2024sweagent, CoderR}, we use a temperature of $0.0$ for deterministic action decoding and input truncation to manage context length. 

\subsection{Experimental Environments}
Our code primarily relies on Python 3.12 and PyTorch 2.7.0. Models are self-supervised fine-tuned with \texttt{LLaMA-Factory}~\citep{zheng2024llamafactory} \footnote{\href{https://github.com/hiyouga/LLaMA-Factory.git}{https://github.com/hiyouga/LLaMA-Factory.git}}, 
and inference is performed with \texttt{vLLM}~\citep{kwon2023efficient}.

\section{Additional Experimental Results}
\label{app:experiments}

\subsection{Additional Analysis of SFT Performance on \ourbench}
\label{app:sft}

\begin{figure}[ht]
    \centering
    \begin{subfigure}{0.48\textwidth}
        \includegraphics[width=\linewidth]{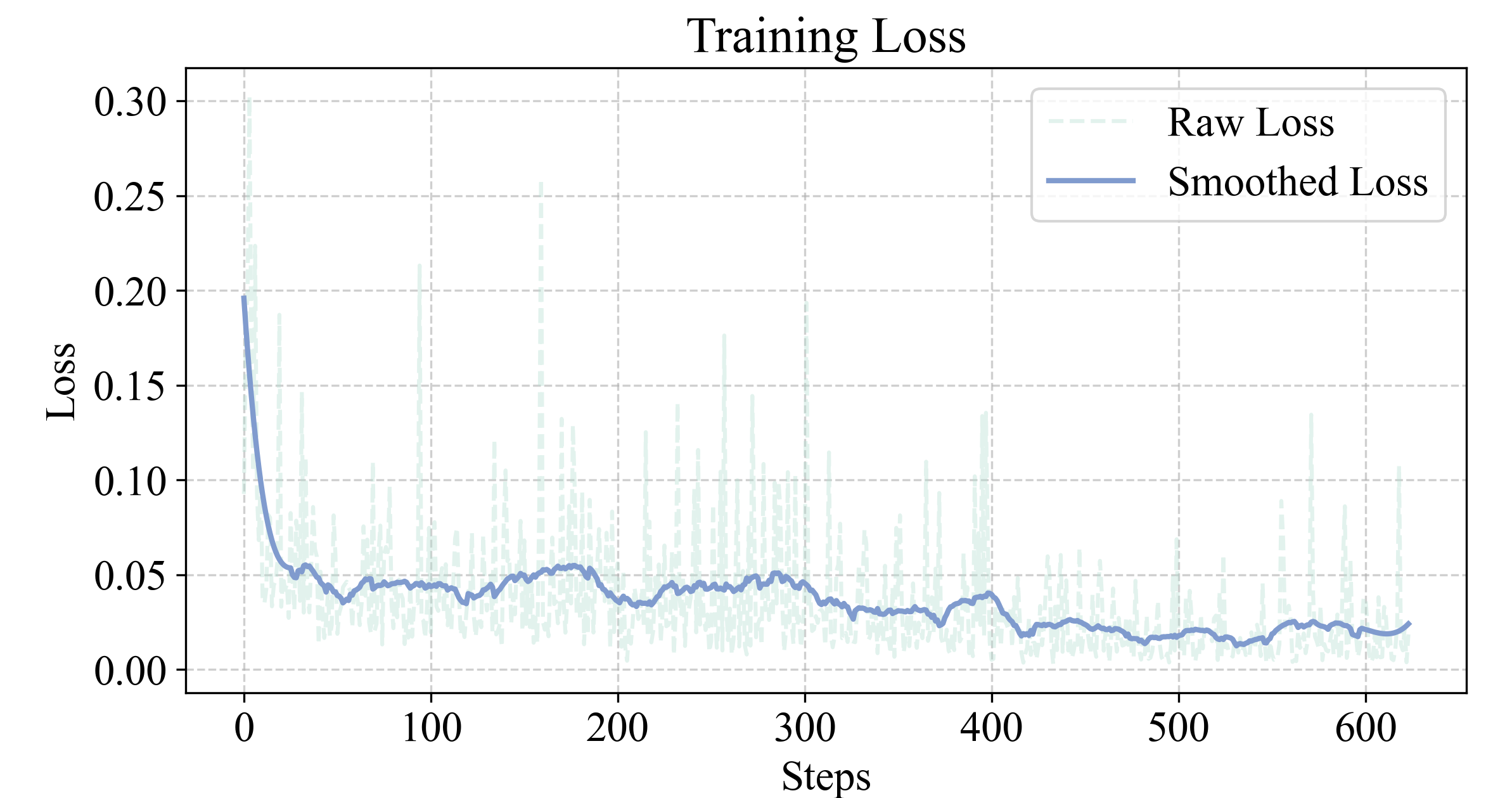}
        \caption{Training loss curve.}
        \label{fig:loss}
    \end{subfigure}
    \hfill
    \begin{subfigure}{0.48\textwidth}
        \includegraphics[width=\linewidth]{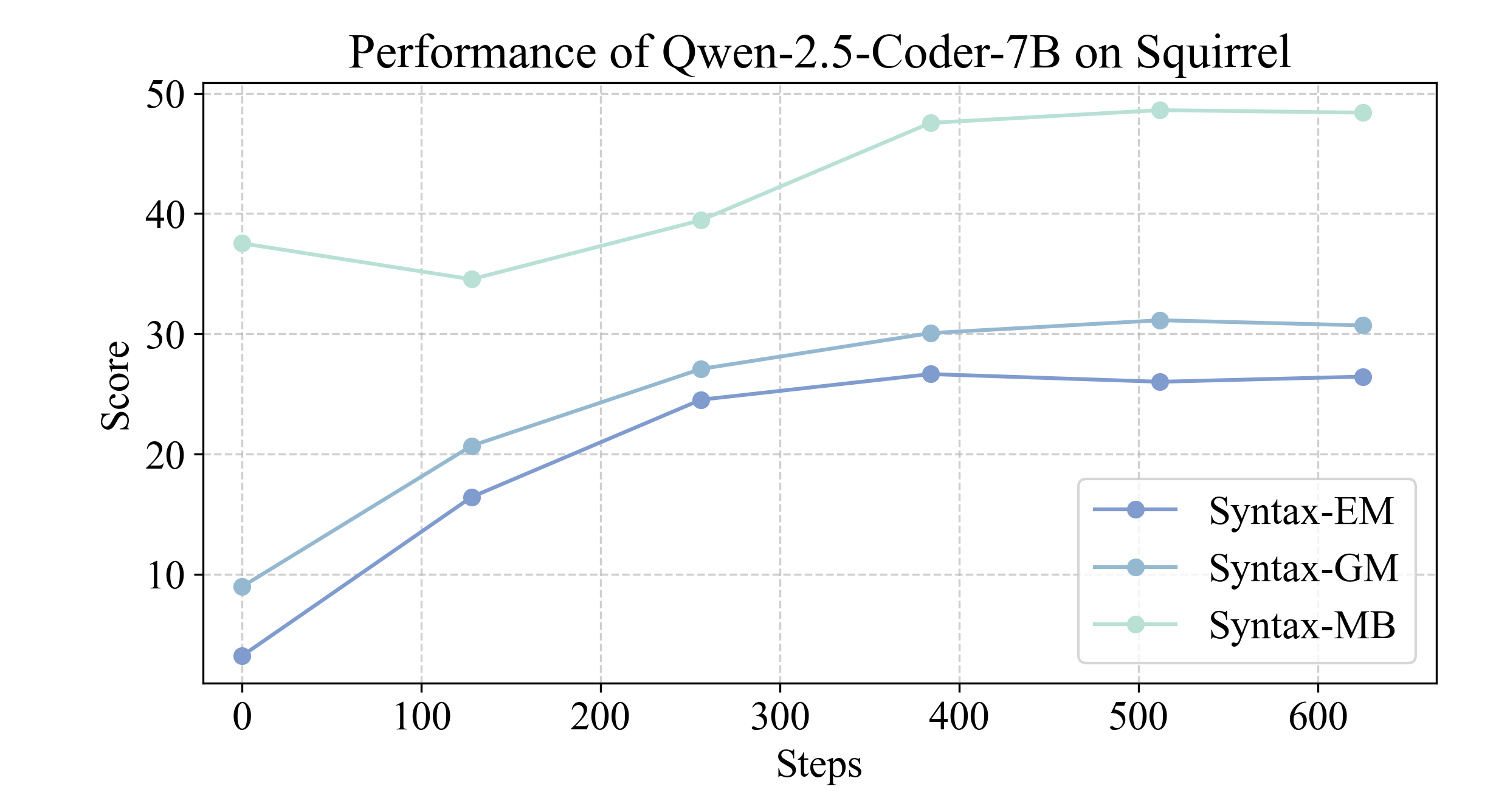}
        \caption{Performance at different training steps.}
        \label{fig:sl}
    \end{subfigure}
    \caption{Analysis of Qwen-2.5-Coder-7B Vanilla SFT on \ourbench, showing corresponding training loss and step-wise performance.}
\end{figure}

\paragraph{Rapid loss decay in SQL debugging fine-tuning.}
Figure~\ref{fig:loss} illustrates that the training loss quickly drops below 0.05 within a few steps, approaching zero. This behavior arises because the constructed SQL debugging parallel data contain inputs with error messages and issue SQL statements, and outputs with the corrected SQL. In most cases, only a small portion of tokens differ between the input and output. Consequently, the model primarily copies tokens from the input, leading to extremely low training loss. When the majority of output tokens carry minimal information, the model tends to ignore the truly informative segments that require correction.

\paragraph{Performance improves with increased training steps.}
Figure~\ref{fig:sl} shows that as training progresses, model performance steadily improves, particularly during the early steps. Beyond approximately 400 training steps, the gains become marginal, indicating diminishing returns. This trend suggests that while additional in-domain training helps, the benefit of further fine-tuning eventually saturates.

\section{Limitations}
\label{app:limitation}

This work introduces a benchmark for enterprise SQL debugging, providing a foundation for future research in software engineering. However, several limitations remain.

\textbf{First, the synthetic nature of the benchmark. }Although the dataset is automatically generated by LLMs, we manually inspected and cross-validated cases that all models failed (detailed in Section \ref{sec: validation}). Nevertheless, undetected artifacts may still exist. Developing more robust automated validation methods is an important direction for future work.

\textbf{Second, constraints of the evaluation framework.} Our rule-based, execution-free evaluation combines exact match, graph match, and edit-direction criteria (detailed in Appendix \ref{app: evaluation metrics}). While effective for debugging scenarios where minimal and precise fixes are expected, this approach is inherently limited by its reliance on reference solutions. For more semantic tasks in which solutions may vary widely, a more flexible and semantics-aware evaluation methodology is needed. We identify this as an area for improvement.

\textbf{Third, the limited SQL dialect coverage.} \ourbench is currently built on Hive/Spark SQL, one of the most widely adopted dialects in large-scale enterprise data infrastructures. Although broader dialect coverage would be valuable, our construction methodology is fundamentally dialect-agnostic, allowing datasets to be synthesized for other SQL dialects. Future iterations will explore additional dialects to expand the benchmark’s coverage.

\clearpage
\section{Case Study}
\label{app:case study}

\begin{figure*}[ht]
    \centering
    \includegraphics[width=1\linewidth]{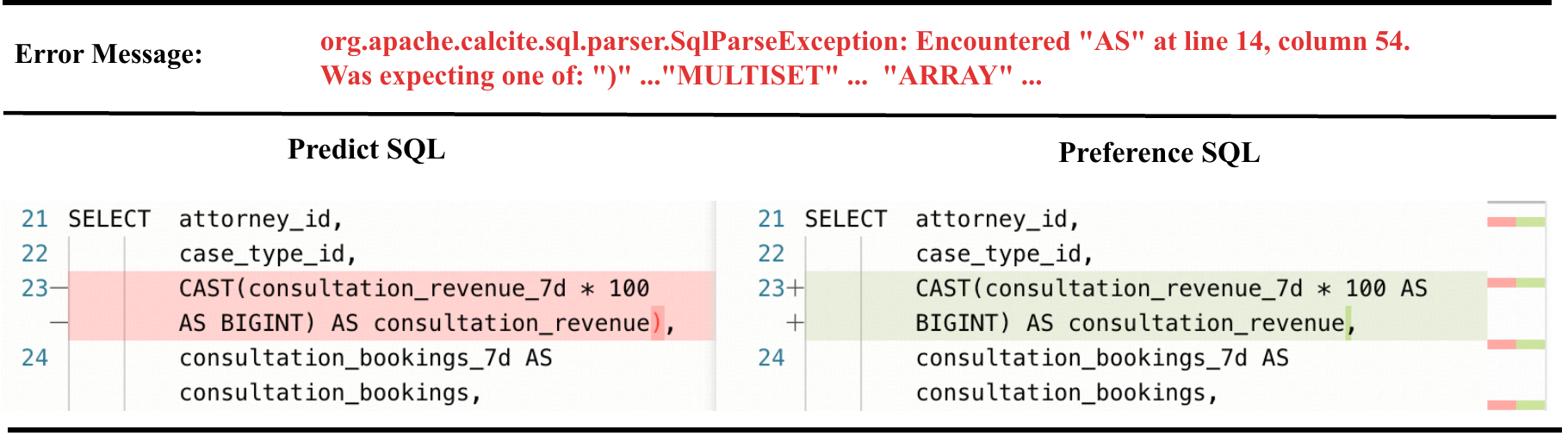}
    \caption{\textbf{Model Hallucination:} After modifying the code according to the error message, the model also inserted an extra ``)'' in similar fragments, which caused the fix to fail.}
    \label{fig:case1}
\end{figure*}

\begin{figure*}[ht]
    \centering
    \includegraphics[width=1\linewidth]{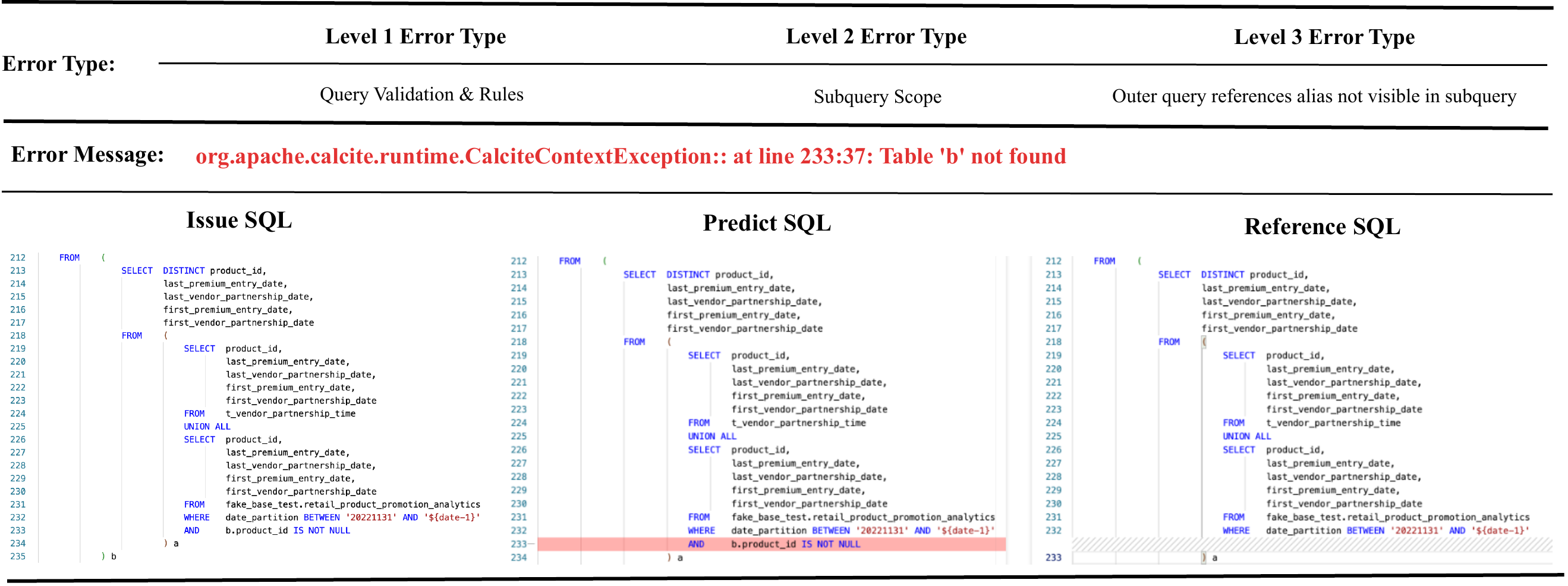}
    \caption{\textbf{Long Context Reasoning Limitation:} The error code uses a non-existent table b (which is usually an alias for a longer table name in SQL), but the model fails to detect this error during the repair process.}
    \label{fig:case2}
\end{figure*}

\section{Prompts Template}
\label{sec:prompts}

The detailed prompts are described below.

\subsection{Enterprise-level SQL Scripts Generation Prompts}

\begin{tcolorbox}[breakable, title={Prompt for Scenario Creation}]
\footnotesize
\label{prompt:mock_ddl}
$\#\#$ Instruction \\

You are a professional SQL ETL and schema generation expert. 
Your task is to transfer a database schema from a source domain to a target domain, preserving structural complexity and table relationships, but fully adapting table names, field names, and semantics to the target domain.
\\

$\#\#$ Steps\\

1. Analyze Source DDL:\\
- Examine the number of tables, fields, data types, relationships, and naming patterns.\\
- Treat this as a structural seed for generating an equivalent schema.\\

2. Generate Target Schema:\\
- Create a logically equivalent schema under the target domain.\\
- Rules:\\
    - Use the database 
    \verb|fake_base_test|.\\
    - Format: CREATE TABLE IF NOT EXISTS \verb|fake_base_test.table_name ( ... )|;\\
    - Avoid SQL reserved keywords as column names.\\
    - Reflect business meaning in the target domain.\\
    - Optionally add auxiliary fields to maintain equivalent complexity.\\
    - All names, comments, and logic must be consistent with the target domain and unrelated to the source domain.\\

3. Validation:\\
- Ensure DDL syntax is correct.\\
- Ensure schema and scenario are fully adapted to the target domain, with no remnants from the source.\\

$\#\#$ Notes\\
- Do not reuse proprietary identifiers or field names from the source domain.\\
- Only use the user-provided target domain.\\
- Preserve the structural pattern, complexity, and relationships of the source schema.\\

$\#\#$ Input Data\\

Source DDL:
\texttt{DDL}\\
Target Domain:
\texttt{SCENARIO}\\

$\#\#$ Output Format(JSON)\\
\begin{verbatim}
{
    "mock scenario": "Scenario description",
    "mock ddl": "Corresponding CREATE TABLE statements"
}
\end{verbatim}
\end{tcolorbox}

\begin{tcolorbox}[breakable, title={Prompt for Generating Enterprise-level SQL}]
\footnotesize
\label{prompt:mock_sql}
$\#\#$ Instruction \\

You are a professional SQL ETL code generation expert.
Using the provided source SQL as a reference, and given the target domain scenario and its corresponding DDL, generate an SQL ETL script for the target domain that preserves the logical structure and complexity of the source code while adapting it fully to the target domain.\\

$\#\#$ Requirement\\

1. Logical structure equivalence:\\
   - Analyze the ETL workflow, table relationships, and processing steps in the source SQL code.\\
   - Preserve the overall structure, complexity, and transformation logic, but replace all table names, field names, and data types to match the target domain.\\

2. Strictly match the target DDL:\\
   - All SQL must be fully based on the provided target DDL.\\
   - Table names and field names must match the target DDL exactly.\\
   - Do not retain any original business terms, identifiers, or domain concepts from the source code.\\

3. Output requirements:\\
   - The code must be executable, and SQL syntax must be correct.\\
   - Maintain a clear hierarchy and readability (include appropriate comments).\\
   - Naming should reflect the target business domain, ensuring a one-to-one correspondence between SQL and the target DDL.\\

$\#\#$ Input Data\\

Source SQL:
\texttt{SQL}\\
Target Domain Scenario:
\texttt{SCENARIO}\\
Target DDL:
\texttt{DDL}\\

$\#\#$ Output Format\\
\begin{verbatim}
{
    'mock code': 'Generated target domain SQL ETL code'
}
\end{verbatim}
\end{tcolorbox}

\subsection{Issue SQL Construction Prompts}
\begin{tcolorbox}[breakable, title={Prompt for Error Type Selection}]
\footnotesize
\label{prompt:error type selection}
$\#\#$ Role:\\
You are an expert SQL engineer specializing in designing realistic SQL bugs for testing and debugging scenarios.  \\

$\#\#$ Task:\\
Given a correct SQL query, your job is to:  \\
Select the top \verb|{TOP_K}| appropriate error type from the provided taxonomy.  \\

$\#\#$Key Guidelines:\\
- Minimal Change: Only introduce the chosen bug. Do not alter the original query’s structure or intent more than necessary.  \\
- Realism: The bug should reflect mistakes that real developers are likely to make. \\  

$\#\#$Input:\\
1. Correct SQL:  
\verb|{SQL} | 

2. DDL (optional):  
\verb|{DDL} | 

3. Original Intent:  
\verb|{CODE INTENTION} | 

4. Error Type Taxonomy:  
\verb|{SEMANTIC ERROR TYPES} | \\

$\#\#$Output Requirements:\\
Your output must include:  \\
- The selected error type(s) at Level 1–3 granularity.  \\

$\#\#$Output Format:
\begin{verbatim}
{
    candidate_errors: 
        {
            "level1_error_type": Level 1 error type,
            "level2_error_type": Level 2 error type,
            "level3_error_type": Level 3 error type
        },
        {
            "level1_error_type": Level 1 error type,
            "level2_error_type": Level 2 error type,
            "level3_error_type": Level 3 error type
        },
    ]
}
\end{verbatim}
\end{tcolorbox}

\begin{tcolorbox}[breakable, title={Prompt for \ourbenchsyn Issue SQL Construction}]
\footnotesize
\label{prompt:issue sql construction}
$\#\#$ Role:\\
You are an expert SQL engineer specializing in designing realistic SQL bugs for testing and debugging scenarios.  \\

$\#\#$ Task:\\
Given a correct SQL query, your task is to introduce an error into the correct query with the smallest possible change.  \\

$\#\#$ Key Guidelines:\\
- Minimal Change: Only introduce the chosen bug. Do not alter the original query’s structure or intent more than necessary.  \\
- Realism: The bug should reflect mistakes that real developers are likely to make.\\  

$\#\#$ Input:\\
1. Correct SQL:  
\verb|{SQL}|  \\

2. DDL (optional):  
\verb|{DDL}|  \\

3. Original Intent:  
\verb|{CODE INTENTION} |   \\

4. Error Type Taxonomy: 
\verb|{SEMANTIC ERROR TYPES}|  \\

$\#\#$ Output Requirements:\\
Your output must include:  \\
- The selected error type(s) at Level 1–3 granularity.  \\
- The modified SQL query with the injected bug.  \\

$\#\#$ Output Format:\\
\begin{verbatim}
{
    "level1_error_type": Level 1 error type,
    "level2_error_type": Level 2 error type,
    "level3_error_type": Level 3 error type,
    "issue_sql": SQL query with the injected bug
}
\end{verbatim}
\end{tcolorbox}

\begin{tcolorbox}[breakable, title={Prompt for \ourbenchsem Issue SQL Construction}]
\footnotesize
\label{prompt:issue sql construction 1}
$\#\#$ Role:\\
You are an expert SQL engineer specializing in designing realistic SQL bugs for testing and debugging scenarios.  \\

$\#\#$ Task:\\
Given a correct SQL query, your job is to:  \\
1. Introduce the error into the SQL query with the smallest possible change.  \\
2. Write a realistic user-style issue report describing how the bug causes the query to behave incorrectly, and the user's real intention.  \\

$\#\#$ Key Guidelines:\\
- Minimal Change: Only introduce the chosen bug. Do not alter the original query’s structure or intent more than necessary.  \\
- Realism: The bug should reflect mistakes that real developers are likely to make.\\  

$\#\#$ Input:\\
1. Correct SQL:  
\verb|{SQL}|  \\

2. DDL (optional):  
\verb|{DDL}|  \\

3. Original Intent:  
\verb|{CODE INTENTION} |   \\

4. Error Type Taxonomy: 
\verb|{SEMANTIC ERROR TYPES}|  \\

$\#\#$ Output Requirements:\\
Your output must include:  \\
- The selected error type(s) at Level 1–3 granularity.  \\
- The modified SQL query with the injected bug.  \\
- A natural-language user bug report describing the mismatch between expected and actual results (without exposing SQL code, since the user does not know the root cause).  \\

$\#\#$ Output Format:\\
\begin{verbatim}
{
    "level1_error_type": Level 1 error type,
    "level2_error_type": Level 2 error type,
    "level3_error_type": Level 3 error type,
    "user_query": Bug report written in natural language. 
    Describe the expected vs. actual outcome clearly.
    "issue_sql": SQL query with the injected bug
}
\end{verbatim}
\end{tcolorbox}

\subsection{Benchmark Evaluation Prompt}

\begin{tcolorbox}[breakable, title={Prompt for \ourbenchsyn Generation}]
\footnotesize
\label{prompt:benchmark evaluation for syntax}
You are an SQL assistant. \\

$\#\#$ Task \\

Based on the error messages and table schema, your task is to fix the issue in the SQL and write the correct SQL. \\
Remember that you can not change any existing comments and SQL code without errors.\\

$\#\#$ Input Data \\
The issue SQL: \texttt{BUG SQL}\\
Related tables schema: \texttt{DDL}\\
Error Messages: \texttt{ERROR MESSAGE}\\

$\#\#$ Output (JSON):\\
\begin{verbatim}
{
    'predict_sql': The fixed SQL.
}
\end{verbatim}
\end{tcolorbox}

\begin{tcolorbox}[breakable, title={Prompt for \ourbenchsem Generation}]
\label{prompt:benchmark evaluation for semantic}
You are an SQL assistant. \\

$\#\#$ Task \\

Based on the user query and input table schema, please fix the bugs in the Issue SQL and write the corresponding correct SQL code.\\
Remember that you can not change any existing comments and SQL code without errors. \\

$\#\#$ Input Data \\
User Query:\texttt{USER QUERY}\\
Related tables schema: \texttt{DDL}\\
Error Messages: \texttt{ERROR MESSAGE}\\

$\#\#$ Output (JSON):\\
\begin{verbatim}
{
    'predict_sql': The fixed SQL.
}
\end{verbatim}
\end{tcolorbox}



\begin{tcolorbox}[breakable, title={Prompt for diff Generation}]
\label{prompt:benchmark evaluation for diff}
\footnotesize
\begin{verbatim}
<background_info>
\texttt{DDL_PLACEHOLDER}
</background_info>

```code
SQL_CODE_PLACEHOLDER
```

<error_msg>
ERROR_MESSAGE_PLACEHOLDER
</error_msg>

```last_edit
<<<<<<< SEARCH
LAST_EDIT_BEFORE_PLACEHOLDER
=======
LAST_EDIT_AFTER_PLACEHOLDER
>>>>>>> REPLACE
```
\end{verbatim}\\
\end{tcolorbox}

\subsection{Agent Prompt}

\begin{tcolorbox}[breakable, title={Prompt for Main Agent}]
\footnotesize
\label{prompt:main agent}

You are a SQL expert. Please review the SQL code (with the table DDL) and the error message reported. Your task is to analyze the error and provide fixing edit instructions.\\

\textbf{Input:}\\
- Tables DDL\\
\verb|DDL_PLACEHOLDER| \\
- Hive SQL Code: \\
\verb|```sql  SQL_CODE_PLACEHOLDER```|\\
- Error Message: \\
\verb|ERROR_MESSAGE_PLACEHOLDER| \\

\textbf{Output Requirements:}\\
You must strictly follow this XML format in your response:\\

\verb|<analysis>|\\
Examine the error message and identify the root cause. Explain what is wrong with the current code and why the error occurred.\\
\verb|</analysis>|\\

\verb|<instructions>|\\
Provide clear, step-by-step instructions on how to fix the code. Explain what changes need to be made and where they should be applied.\\
\verb|</instructions>|\\

\verb|<sketch_sql>|\\
Provide the edit sketch using the special comment \verb|`...`| to represent unchanged code between edited lines. Specify each edit in sequence, minimizing unchanged SQL code while making it clear what the edit is and where it should be applied.\\
\verb|</sketch_sql>|\\

Ensure your instructions(in Chinese) and sketch are clear enough that another model can apply them correctly without accidentally deleting or modifying unintended parts of the code.
\end{tcolorbox}

\clearpage
\section{SQL Bug Taxonomy}
\label{app:SQL Bug Taxonomy}
\subsection{Bug Distribution of \ourbenchsem}
\begin{table}[ht]
\small
\setstretch{1.5}
\centering
\caption{Error type distribution in \ourbenchsem}
\label{tab:semantic}
\resizebox{\textwidth}{!}{%
\begin{tabular}{l|l|l|c}
\toprule[1.2pt]
\rowcolor[HTML]{ECF4FF} 
\multicolumn{1}{c|}{\cellcolor[HTML]{ECF4FF}\textbf{Level 1 Error Type}} &
  \multicolumn{1}{c|}{\cellcolor[HTML]{ECF4FF}\textbf{Level 2 Error Type}} &
  \multicolumn{1}{c|}{\cellcolor[HTML]{ECF4FF}\textbf{Level 3 Error Type}} &
  \textbf{Count} \\ \midrule[1.2pt]
 &
   &
  Using COUNT(column) instead of COUNT(*) and misunderstanding NULL exclusion &
  43 \\
 &
  \multirow{-2}{*}{Aggregate Logic} &
  Using SUM()/AVG() on a column with NULLs without COALESCE &
  27 \\ \cline{2-4} 
 &
   &
  JOIN condition placed in WHERE clause (accidental CROSS JOIN) &
  41 \\
 &
   &
  Failing to handle NULLs in JOIN keys (causing rows to disappear) &
  13 \\
 &
  \multirow{-3}{*}{Join Logic} &
  Missing condition causing Cartesian product &
  2 \\ \cline{2-4} 
 &
   &
  Three-valued logic error: NOT (a = b) not equivalent to a != b when NULLs present &
  14 \\
 &
  \multirow{-2}{*}{Boolean \& Logic} &
  Improper Boolean usage (e.g., WHERE col = TRUE) &
  9 \\ \cline{2-4} 
 &
   &
  NULL compared with = (should use IS NULL) &
  33 \\
 &
  \multirow{-2}{*}{NULL Handling} &
  Confusion between IS NULL and =NULL &
  2 \\ \cline{2-4} 
 &
   &
  Using RANK() instead of ROW\_NUMBER() or DENSE\_RANK() leading to duplicates/skips &
  31 \\
 &
  \multirow{-2}{*}{Window Function Logic} &
  Incorrect partitioning/ordering in window function leading to wrong row assignment &
  4 \\ \cline{2-4} 
 &
   &
  Misplaced LIMIT inside subquery affecting outer results &
  3 \\
 &
  \multirow{-2}{*}{Subquery Scope} &
  Correlated subquery missing correlation condition &
  2 \\ \cline{2-4} 
 &
   &
  Missing condition causing Cartesian product &
  12 \\
 &
  \multirow{-2}{*}{JOIN Logic} &
  Wrong join key used inside nested subquery &
  2 \\ \cline{2-4} 
 &
  Set Operations &
  UNION vs. UNION ALL misuse (unintended deduplication) &
  55 \\ \cline{2-4} 
 &
  Date/Time Logic &
  Confusion between DATE, TIMESTAMP, and INTERVAL types &
  23 \\ \cline{2-4} 
\multirow{-18}{*}{Semantics \& Logic} &
  Pattern Matching &
  Incorrect LIKE usage &
  2 \\ \hline
 &
  Separator Rule &
  collect\_set/concat\_ws separator uses semicolon &
  54 \\ \cline{2-4} 
\multirow{-2}{*}{Functions \& Expressions} &
  Function Semantics &
  Misunderstanding the empty handling of aggregate functions &
  1 \\ \hline
 &
   &
  Misuse of ROLLUP / CUBE &
  14 \\
 &
  \multirow{-2}{*}{GROUP BY Extensions} &
  Rollup/Cube/Grouping Sets producing unexpected super-aggregate rows &
  3 \\ \cline{2-4} 
 &
   &
  Grouping by a functionally dependent column unnecessarily &
  17 \\
 &
  \multirow{-2}{*}{GROUP BY Logic} &
  Rollup/Cube/Grouping Sets producing unexpected super-aggregate rows &
  4 \\ \cline{2-4} 
\multirow{-5}{*}{Joins \& Grouping} &
  JOIN Type Selection &
  Using INNER JOIN when LEFT JOIN is needed (loss of data) &
  64 \\ \hline
 &
   &
  Duplicate rows due to many-to-many join not being accounted for &
  1 \\
\multirow{-2}{*}{Result \& Quality} &
  \multirow{-2}{*}{Correctness} &
  Incorrect output data &
  1 \\ \hline
 &
  Implicit Casting &
  Implicit cast changing semantics (e.g., string to number) &
  15 \\ \cline{2-4} 
\multirow{-2}{*}{Types \& Data Formats} &
  Data Format &
  Misused format placeholder &
  1 \\ \hline
Identifiers \& Objects &
  Qualification &
  Qualifying a column with the wrong table alias in a complex join &
  22 \\ \bottomrule[1.2pt]
\end{tabular}%
}
\end{table}

\subsection{Bug Distribution of \ourbenchsyn}
\begin{table}[ht]
\footnotesize
\setstretch{0.9}
\centering
\caption{Error type distribution in \ourbenchsyn}
\label{tab:syntax}
\resizebox{\textwidth}{!}{%
\begin{tabular}{l|l|l|c}
\toprule[1.2pt]
\rowcolor[HTML]{ECF4FF} 
\multicolumn{1}{c|}{\cellcolor[HTML]{ECF4FF}\textbf{Level 1 Error Type}} &
  \multicolumn{1}{c|}{\cellcolor[HTML]{ECF4FF}\textbf{Level 2 Error Type}} &
  \multicolumn{1}{c|}{\cellcolor[HTML]{ECF4FF}\textbf{Level 3 Error Type}} &
  \textbf{Count} \\ \midrule
                                             &                                          & Missing parameter for explode                                  & 15 \\
                                             &                                          & Incorrect explode parameter                                    & 9  \\
                                             &                                          & explode(map) requires two aliases                              & 2  \\
                                             &                                          & date\_add missing parameter (also typo data\_add)              & 2  \\
                                             & \multirow{-5}{*}{Parameter Completeness} & array\_contains wrong argument type                            & 1  \\ \cline{2-4} 
                                             &                                          & get\_json\_object wrong argument type                          & 4  \\
                                             &                                          & array\_contains wrong argument type                            & 3  \\
                                             & \multirow{-3}{*}{Parameter Type}         & from\_json wrong argument type                                 & 1  \\ \cline{2-4} 
                                             &                                          & Missing LATERAL VIEW                                           & 94 \\
                                             & \multirow{-2}{*}{LATERAL VIEW Required}  & Missing alias for LATERAL VIEW function output                 & 1  \\ \cline{2-4} 
                                             & Date Difference                          & datadiff argument/typo error                                   & 5  \\ \cline{2-4} 
                                             & Type Conversion                          & Multiple AS in CAST                                            & 15 \\ \cline{2-4} 
                                             & Nesting Limit                            & Aggregate expressions cannot be nested                         & 2  \\ \cline{2-4} 
                                             & Separator Rule                           & collect\_set/concat\_ws separator uses semicolon               & 11 \\ \cline{2-4} 
                                             & Date/Time                                & to\_unix\_timestap typo                                        & 1  \\ \cline{2-4} 
\multirow{-16}{*}{Functions \& Expressions}  & Function Spelling                        & concat\_ws typo                                                & 1  \\ \hline
                                             &                                          & Missing END or THRN in CASE WHEN                               & 72 \\
                                             & \multirow{-2}{*}{CASE Expression}        & Multiple END in CASE WHEN                                      & 4  \\ \cline{2-4} 
                                             &                                          & Missing argument in IN                                         & 7  \\
                                             & \multirow{-2}{*}{Conditional Logic}      & IN subquery returns multiple columns                           & 1  \\ \cline{2-4} 
                                             &                                          & Window function misused with GROUP BY                          & 3  \\
                                             &                                          & Window function used inside WHERE/HAVING                       & 3  \\
                                             & \multirow{-3}{*}{Window Functions}       & Window function frame clause misuse (e.g., ROWS BETWEEN error) & 1  \\ \cline{2-4} 
                                             & Subquery Scope                           & Outer query references alias not visible in subquery           & 2  \\ \cline{2-4} 
                                             & Aggregation \& Subquery                  & SELECT list contains non-aggregated column not in GROUP BY     & 62 \\ \cline{2-4} 
                                             & Pattern Matching                         & Incorrect LIKE usage                                           & 2  \\ \cline{2-4} 
                                             & Aggregate Usage                          & Aggregate function in SELECT without GROUP BY                  & 1  \\ \cline{2-4} 
\multirow{-12}{*}{Query Validation \& Rules} & Boolean \& NULL                          & NULL compared with = (should use IS NULL)                      & 2  \\ \hline
                                             &                                          & Incorrect clause ordering - JOIN after WHERE                   & 7  \\
                                             &                                          & Invalid SELECT clause syntax with subquery                     & 6  \\
                                             &                                          & Missing SELECT before FROM clause                              & 5  \\
                                             &                                          & Multiple WHERE                                                 & 4  \\
                                             &                                          & Missing partition conditions in WHERE clause                   & 3  \\
                                             &                                          & Missing logical connector in WHERE                             & 26 \\
                                             &                                          & Non-query expression in illegal context                        & 3  \\
                                             &                                          & Missing FROM clause                                            & 2  \\
                                             & \multirow{-9}{*}{Clause Structure}       & Column count mismatch in UNION                                 & 1  \\ \cline{2-4} 
                                             &                                          & WITH AS not first                                              & 26 \\
                                             &                                          & Unnecessary WITH AS                                            & 13 \\
                                             & \multirow{-3}{*}{CTE/View}               & Trailing comma after last view                                 & 23 \\ \cline{2-4} 
                                             &                                          & Keyword spelling error                                         & 3  \\
                                             &                                          & Space in !=                                                    & 2  \\
                                             & \multirow{-3}{*}{Keywords \& Operators}  & Missing IN keyword                                             & 2  \\ \cline{2-4} 
                                             & Statement Ending                         & Extra trailing statements                                      & 4  \\ \cline{2-4} 
                                             & Parentheses / Brackets                   & Missing closing parenthesis                                    & 5  \\ \cline{2-4} 
                                             & Alias / AS                               & Redundant AS                                                   & 3  \\ \cline{2-4} 
\multirow{-19}{*}{Grammar \& Structure}      & SELECT List                              & Missing column list after SELECT                               & 1  \\ \hline
                                             &                                          & Variable error                                                 & 13 \\
                                             &                                          & Missing partition conditions in DELETE statement               & 2  \\
                                             & \multirow{-3}{*}{Variables/Placeholders} & Partition column comparison with numeric type not allowed      & 2  \\ \cline{2-4} 
                                             &                                          & Column exists in multiple tables but alias omitted             & 8  \\
                                             & \multirow{-2}{*}{Ambiguous References}   & Ambiguous alias in nested subquery with same column name       & 1  \\ \cline{2-4} 
                                             &                                          & Field/Table does not exist                                     & 11 \\
                                             & \multirow{-2}{*}{Schema/Object}          & Missing partition query conditions                             & 2  \\ \cline{2-4} 
\multirow{-8}{*}{Identifiers \& Objects}     & Naming/Alias                             & Duplicate names (column/alias)                                 & 5  \\ \hline
                                             &                                          & Missing grouping column                                        & 14 \\
                                             & \multirow{-2}{*}{GROUP BY}               & Missing HAVING clause for aggregate filtering                  & 1  \\ \cline{2-4} 
                                             &                                          & Missing condition causing Cartesian product                    & 6  \\
                                             & \multirow{-2}{*}{JOIN Ambiguity}         & Missing table prefix for duplicate column names in join        & 35 \\ \cline{2-4} 
\multirow{-5}{*}{Joins \& Grouping} &
  Nested Joins &
  Ambiguous column reference due to multiple levels of alias &
  1 \\ \hline
                                             &                                          & Punctuation error                                              & 49 \\
                                             &                                          & Incorrect quote type for column alias with special characters  & 4  \\
\multirow{-3}{*}{Punctuation \& Formatting} &
  \multirow{-3}{*}{Punctuation/Parentheses} &
  Missing semicolon between statements &
  5 \\ \hline
                                             &                                          & Insert error                                                   & 37 \\
                                             & \multirow{-2}{*}{Insert Statement}       & Mismatched column count                                        & 5  \\ \cline{2-4} 
\multirow{-3}{*}{DML \& DDL}                 & Create Table Statement                   & Table creation error                                           & 10 \\ \hline
                                             &                                          & TRANSFORM with lambda expression not supported in Hive         & 3  \\
\multirow{-2}{*}{Compatibility/Dialect} &
  \multirow{-2}{*}{Function Differences} &
  wm\_concat function not supported in the current SQL dialect &
  1 \\ \hline
                                             & Type System                              & Type mismatch                                                  & 16 \\
\multirow{-2}{*}{Types \& Data Formats}      & Date/Time                                & to\_unix\_timestap typo                                        & 2  \\ \bottomrule[1.2pt]
\end{tabular}%
}
\end{table}

\section{Examples}
\label{sec:example}
\begin{figure}[th]
    \centering
    \includegraphics[width=0.9\linewidth]{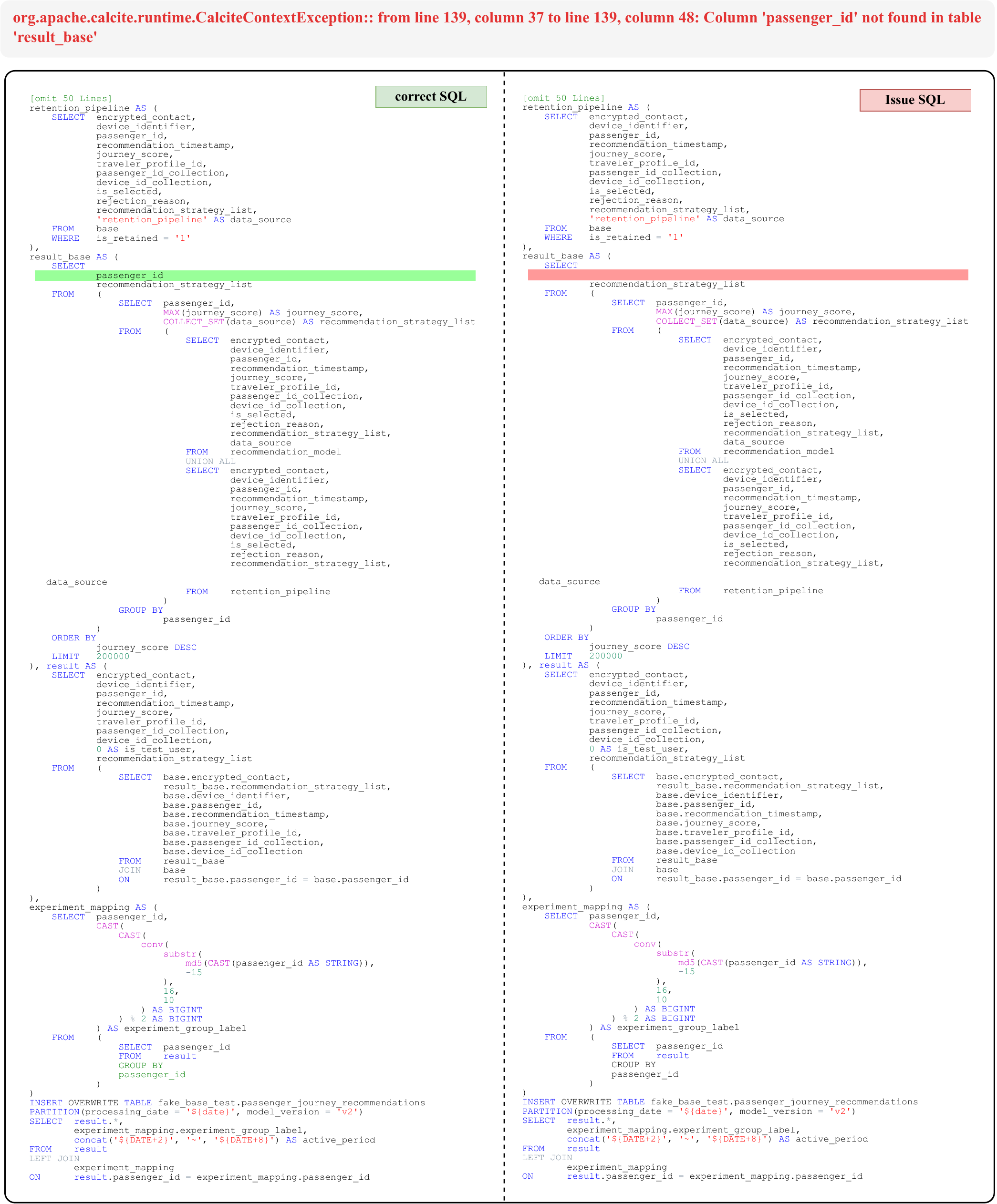}
    \caption{The example of \ourbenchsyn, where an explicit error message exists.}
    \label{fig:syntax_example}
\end{figure}

\begin{figure}[t]
    \centering
    \includegraphics[width=0.9\linewidth]{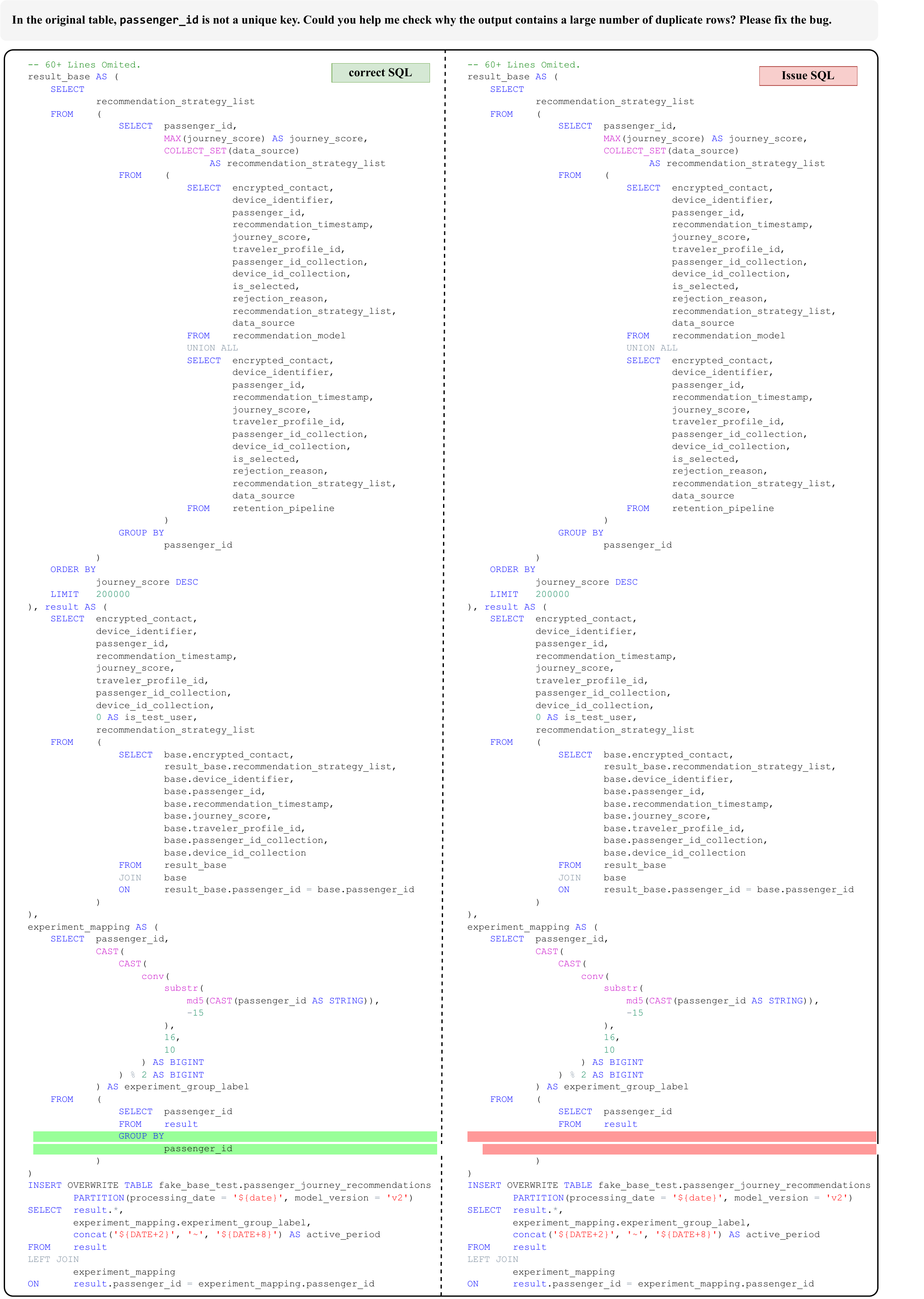}
    \caption{The example of \ourbenchsem.}
    \label{fig:semantic_example}
\end{figure}

\clearpage

\end{document}